# A New Inference algorithm of Dynamic Uncertain Causality Graph Based on Conditional Sampling Method for Complex Cases

Hao Nie and Qin Zhang ∗

∗Corresponding author: qinzhang@tsinghua.edu.cn

## Abstract

Dynamic Uncertain Causality Graph (DUCG) is a recently developed model for fault diagnoses of industrial systems and general clinical diagnoses. In some cases, however, the variable state combination explosion is a problem that may result in the inefficiency or even disability of DUCG inference, i.e., when many intermediate variables are state-unknown while the consequence variables are state-known in a DUCG graph, the state combination explosion of intermediate variables may appear during the inference computation. Monte Carlo sampling is a typical algorithm to solve this type of problem. However, we are facing that the calculation values are very small, e.g. $10^{-15}$, which means a huge number of samplings are needed. This paper proposes an algorithm based on conditional stochastic simulation, which obtains the final calculation result from the expectation of the conditional probability in sampling loops instead of counting the sampling frequency. Compared with the early presented recursive algorithm of DUCG inference, the proposed algorithm requires much less computation time in the case when state-unknown intermediate variables are many. A simple example is given to illustrate the proposed method, which shows that the proposed algorithm reduces the computation time significantly. An example for diagnosing Viral Hepatitis B shows that the new algorithm performs 3 times faster than the recursive algorithm with 2.7% error ratio. The effect is more when state-unknown variables are more.

## 1. Introduction

Computer-aided diagnosis for finding root causes of system abnormalities are desired for large and complex systems, e.g. fault diagnoses of nuclear power plants and clinical diagnoses. As one of the models, DUCG (dynamic uncertain causality graph) is recently developed [1]–[6]. It is based on domain knowledge with uncertainty and has high diagnosis precision, strong interpretability and without generalization problem that machine learning approaches usually have.

DUCG is developed from dynamic causality diagram (DCD)[7] , formally presented in [1], and keeps evolving in recent years. In [1], the support to incomplete knowledge representation of DUCG was addressed. In [2], the statistic base of DUCG was addressed which shows that the parameters of DUCG can be either learnt from data or specified by domain experts. In[3], algorithm for breaking the directed cyclic graph (DCG) in DUCG was presented. In [8], a new algorithm called cubic-DUCG was introduced for temporal inference of DUCG. In [9], some new type variables were introduced to meet the requirement of practice.

DUCG diagnosis has been applied in many real cases. In [10], a DUCG model is built for the online fault diagnoses of a nuclear power plant's generator system, with which most inference process can be completed within a second just on a laptop computer, even the model is composed of 633 variables and 2952 links between child and parent variables. In [11], a DUCG model with 32 variables and 71 conditional and unconditional causalities are used to monitor the power supply system of a space craft. In [6], DUCG plays the main role in a distributed and web-based system to provide a platform for general clinical diagnosis. The diagnoses were verified by several third-party hospitals.

DUCG is composed of single valued DUCG (*S*-DUCG) and multi-valued DUCG (*M*-DUCG).The so called single-valued means that only the causes of the true state of a child variable can be specified, while the false state can only be the complement of the true state. In contrast, the so called multi-valued means that the causes of all states of a child node can be specified separately. Since *M*-DUCG does not have the limitation of *S*-DUCG, *M*-DUCG is applied in more practical cases. For simplicity, this paper focuses on only *M*-DUCG and M-DUCG is abbreviated as DUCG in the rest of the paper.

The basic idea of DUCG is to represent the uncertain causalities between a child variable and its parent variables by introducing virtual functional random events and causal relationship intensities.

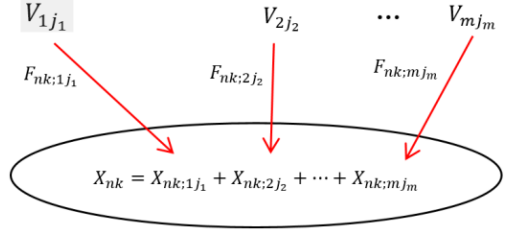

Figure 1: The basic idea of *M*-DUCG model[1]

Figure 1, Eqn.(1)(2) explain the core idea of DUCG model. *V* represents the parent events. The $i$ in $V_{ij_i}$ or briefly $V_{ij}$ indexes the event variable and $j_i$ or briefly $j$ indexes the state of the variable. The lower-case letters represent the probabilities of the corresponding upper-case letters denoting events.

$$X_{nk} = \sum_i \sum_{j_i} X_{nk;ij_i}$$
$$= \sum_i \sum_{j_i} F_{nk;ij_i} V_{ij_i} \quad (1)$$
$$= \sum_i \sum_j (r_{n;i}/r_n) A_{nk;ij} V_{ij}$$

$$x_{nk} = \sum_i \sum_j (r_{n;i}/r_n) a_{nk;ij} v_{ij} \quad (2)$$

In which, $r_{n;i} > 0$ is defined as the causal relationship intensity between $X_n$ and $V_i$, $r_n \equiv \sum_i r_{n;i}$. For simplicity, $j_i$ is simplified as $j$ in the last expression of Eqn.(1) and in the rest of this paper. Each evidence event $X_{nk}$ observed can be expanded to its parent events according to Eqn.(1), and its parents $V_{ij}$ can be further expanded as their parents in the same way. The expanding process will not stop until root causes B or D. Where *B*-type event means a basic/root event, whose probability is given during graph construction and *D*-type event is defined as a default or unknown cause event with probability equal to 1. Then the evidence $E = \prod_{X_{nk} \in E} X_{nk}$ can be expanded as a sum-of-products composed of only {*B*-, *D*-, *A*-}-type events and *r*-type parameters, in which the logic absorption, exclusion and *r*-type parameter calculation are needed. During the expanding, once a state-unknown variable is as a parent, all its states should be expanded leading to a small sum-of–products. More state-unknown variables will cause the multiplication of a set of small sum-of-products and then the state combination exploration will occur.

After obtaining the sum-of-products, the weighted OR operation applies as shown in Eqn.(2), i.e., by replacing the upper-case letters in the sum-of-products in terms of events with the corresponding lower-case letters in terms of probabilities of the events, the probability of *E* can be calculated. Note that the {*b*-, *a*-, *r*-}-type parameters are encoded in the constructed DUCG. The DUCG inference is to calculate $Pr\{B_{kj} | E\} = Pr\{B_{kj}E\}/Pr\{E\}$ for all possible $B_{kj}$ in concern. Details of DUCG are described in Sec.2.1.

It is the same as Bayesian network[12] that the inference problem for DUCG is an NP-hard problem. As addressed above and in Sec.2.3 and Sec.2.4 in details, the expanding work to obtain the final sum-of-products is exponential to the number of nodes and connections, and the combination explosion will make the problem unsolvable when the number of state-unknown intermediate variables increase or the structure of graph become complex. When there are no or only a few state-unknown intermediate variables, recursive algorithm introduced in [5] can be applied to accelerate the inference process. However, when the state-unknown intermediate variables are more than a few, the computation efficiency is a big problem.

The state-unknown intermediate variable can be often encountered in practice. For example, domain expert represents the causality that the failure of dry wet separator may lead to a leak of the valve in a pipeline as shown in Figure 2a , where the red color indicates state determined. The knowledge inside is that the water droplets in the air flow hit the valve which speeds up the corrosion that causes the leak of the valve. Thus, the intermediate event $X_{3,1}$ representing the abnormal high content of water droplets in the airflow should be added to describe this mechanism in details as shown in Figure 2b. Since $X_3$ is not monitored by any sensor ($X_{3,0}$ indicates normal content of water droplets), it can be regarded as not observable and thus state-unknown. The DUCG graph is as shown in Figure 2c, in which $X_3$ without color is a state-unknown intermediate variable.

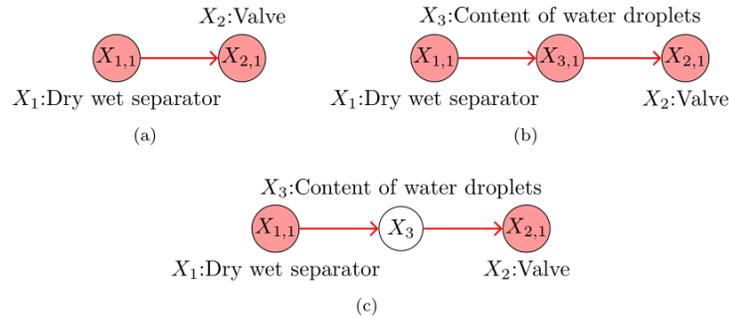

Figure 2: A simple example of state-unknown variable in DUCG

Figure 3 describes another example[6] for a patient who suffers a nasal septum deviation. His case record is: A middle-aged male with no history of trauma, unilateral nasal congestion, persistent nasal obstruction, nasal itching, unilateral epistaxis, volume of nasal bleeding was less, deviation of nasal septum found by physical examination, other symptoms and physical signs were normal, no laboratory examination and imaging examination results provided.

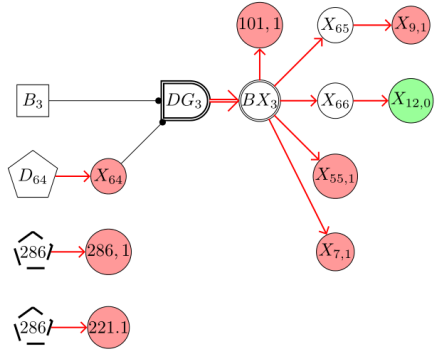

Figure 3: State-unknown intermediate variables in a nasal septum deviation case [6]

The definitions of nodes in Figure 3 are listed in Table 1. In this graph, ⓧ represents for the consequence or effect variables, and $B_3$ represents for the nasal septum deviation. For simplicity, the node shapes represent the types of variable, e.g., (101,1) represents *X*-type variable $X_{101,1}$. Nodes with other shapes DG, D, BX represent for {*DG-, D-, BX-*}-type variables (definitions are given in [6] but not given here because they are not used in the rest of this paper).

| Variable | Variable description |
|---|---|
| $B_3$ | Nasal septum deviation |
| $X_7$ | Nasal obstruction |
| $X_9$ | Hemorrhinia |
| $X_{12}$ | Headache |
| $X_{55}$ | Nasal septum bending (physical examination) |
| $X_{64}$ | History of external head injury |
| $X_{65}$ | Nasal mucosal erosion |
| $X_{66}$ | Partial compression of ipsilateral turbinate |
| $X_{99}$ | Progressive nasal congestion |
| $X_{101}$ | Volume of nasal bleeding |
| $X_{221}$ | Itchy nose |
| $X_{286}$ | State of Stuffy nose |

Table 1: $\{B-, X-\}-type$ variables in Fig.3

In Fig. 3, nodes with red color such as $X_{101}$, $X_{55,1}$, $X_{7,1}$, $X_{9,1}$ are observed abnormal while $X_{12,0}$ with green color means state normal. $X_{66}$ and $X_{65}$ without color are state-unknown variables.

To avoid the state-unknown intermediate variables, a temporary solution is to link the state-known variables directly. But this will cause the difficulty of representation and the interpretability. To represent the detailed knowledge and have more interpretability, the state-unknown intermediate variables are unavoidable, resulting in the state combination explosion. Therefore, how to make an efficient DUCG inference is a problem to be solved.

Monte Carlo method is a common numerical solution for the combination explosion problem. Logic sampling[13], Gibbs sampling[14] and importance sampling[15] have been introduced for BNs inference and perform well. These sampling methods calculate the probability based on the frequencies of events in loops, which is reasonable for BNs as the result is about $10^{-6}$ and about $N = \frac{4}{\phi \epsilon^2} ln \frac{2}{\delta}$ loops are required[16]. $\phi$ in expression is the exact solution, $\epsilon$ is the error ratio of estimate and the estimate result $\mu$ need to meet $P[\phi(1-\epsilon) \leq \mu \leq \phi(1+\epsilon)] \geq 1-\Delta$. However, $\phi$ for DUCG cases could be less than $10^{-15}$ in many practical applications, and the loop number will be too many for calculation.

To overcome this problem, a new DUCG sampling algorithm called conditional stochastic simulation based on the recursive inference algorithm of DUCG[5] is presented in this paper.

In Sec.2, as a background of this paper, more detailed DUCG model and the recursive inference algorithm are briefly introduced. A detailed analysis on the combination explosion problem of DUCG is also proposed. In Sec.3, the detailed scheme for sampling algorithm are presented, and the performance of the sampling method is also addressed in this section. Two ideal models are shown in Sec.4 to verify the accuracy and performance of sampling algorithm. Then, a real model of Viral hepatitis B is given for further comparison. The conclusion is drawn in Sec.5.

## 2. Background Technology

### 2.1. More detailed Introduction to DUCG Model

Similar to the other probabilistic graphical models (PGMs), DUCG is composed of nodes and directed arcs. Node events in DUCG are denoted with different shapes, the detailed definitions are listed in Table 2. For all node events, the first subscript indexes the variable and the second indexes the state.

| Type name | Standard Shape | Simplified Shape | Definition |
|---|---|---|---|
| B-type | $B_{ij}$ | ij | The root cause event |
| X-type | $X_{nk}$ | nk | The consequence or effect variable |
| G-type | $G_i$ | i | The logic gate, has a attached table to show the logic conditions. |
| D-type | $D_n$ | n | the default or unclear parent variables of variable $X_n$. |

Table 2: Shapes of some nodes in DUCG

In DUCG, red directed arcs are regarded as event matrices, which are named as weighted functional events and indicate the causality relationships among nodes, where weight is $(r_{n;i}/r_n)$ and $A_{nk;ij}$ is the functional event, which is a member of event matrix $A_{n;i}$, where $n$ indexes the child variable and $i$ indexes the parent variable. The directed red arc ⟶ is written in text as $F_n = (r_{n;i}/r_n)A_{n;i}$, $a_{nk;ij}=\Pr\{A_{nk;ij}\}$. Definitions of some directed arcs used in DUCG are shown in Table 3.

| Standard Shape | Definition |
|---|---|
| → | Event matrix $F_{n;k}$ |
| -→ | Conditional $F_{n;k}$ with condition event $Z_{n;k}$ |
| —• | Input for a logic gate. |

Table 3: Shape of directed arcs in DUCG

The query for inference of DUCG is to get the probability for each state of possible root causes conditional on evidence $E$, i.e. $Pr\{B_{ij} | E\}$. In Figure 3, $E = X_{7,1}X_{9,1}X_{55,1}X_{101,1}X_{221,1}X_{286,1}$. The basic inference algorithm of DUCG can be roughly summarized as the following steps:

1. Simplify the DUCG according to the observed evidence $E$ according to rules presented in [1]. The B-type events remained are the possible root causes included in $S_H$;

2. Expand $B_{kj}E$ and $E$ with Eqn.(1) as sum-of-product expressions composed of events $B_{ij}$, $A_{nk;ij}$ and $r$-type parameters, where $B_{kj} \in S_H$.

3. When there is more than one member in $S_H$, calculate the numerical values of

$$Pr\{B_{kj} | E\} = \frac{Pr\{B_{kj}E\}}{Pr\{E\}} \quad for \quad all \quad B_{kj} \in S_H \quad (3)$$

Figure 4 shows an example of a 5-layer DUCG which has two $B-type$ variables and three evidences: $X_{17,1}$, $X_{18,1}$ and $X_{19,1}$.

To calculate Eqn.(3), we need to expand $E$, i.e. we need to expand every member in $E$ and multiply them together. For example, evidence $X_{17,1}$ can be expanded with Eqn.(1) as follows:

$X_{17,1} = F_{17,1;13}X_{13} + F_{17,1;13}X_{14} + F_{17,1;13}X_{15} + F_{17,1;13}X_{16}$, and then, $X_{13}$, $X_{14}$, $X_{15}$ and $X_{16}$ can be further expanded to a sum of 4 terms respectively, all including $X_9, X_{10}, X_{11}, X_{12}$. Similarly, $X_9$ through $X_{12}$ can be expanded respectively. For example, $X_9 = F_{9;5}X_5 + F_{9;6}X_6 + F_{9;7}X_7 + F_{9;8}X_8$. Note that $F_{17,1;13}$ through $F_{17,1;15}$ and $F_{9;5}$ through $F_{9;8}$ are event matrices, not single events like $F_{17,1;13,1}$. Similarly, $X_5$ through $X_8$ can be expanded to $X_1$ through $X_4$, and $X_1$ through $X_4$ can be expanded to $B_1$ and $B_2$. It is easy to understand that the final expanded sum-of-products of $E=X_{17,1}X_{18,1}X_{19,1}$ have $4^3 \times 4^4 \times 4^4 \times 4^4 \times 2^4 = 17,179,869,184$ items, without considering the members of event matrices.

In many cases, evidences are in the furthest position from root causes, so that the expanded sum-of-product for each evidence is big. When we multiply them together in expanding $E$, the combination explosion problem appears. More details are addressed in Sec.2.3 During the expanding, a lot of event absorptions and inclusions along with $r$-type parameter calculations are involved. This problem is also part of the combination explosion and is called the logic operation problem in DUCG. More details are addressed in Sec.2.4.

Due to the two problems, it is hard to get the value of $Pr\{E\}$ and $Pr\{B_{kj}E\}$ directly.

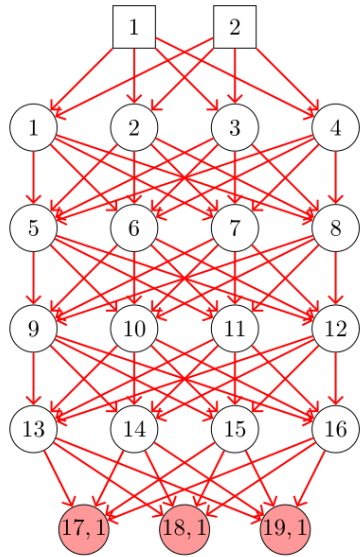

Figure 4: A 5 layers DUCG with 16 state-unknown variables

## 2.2. Recursive Algorithm

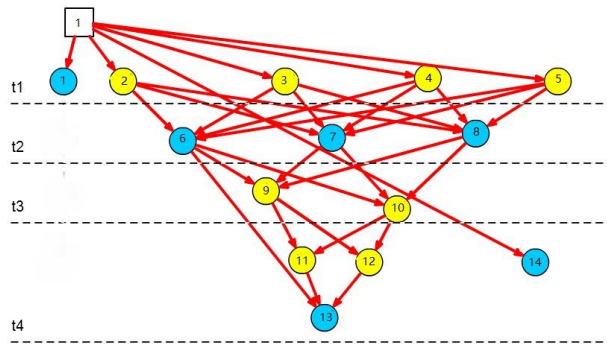

Figure 5: An example for cubic-DUCG with multiple layers.

The recursive algorithm is proposed for the inference of cubic-DUCG[8] in [17]. A typical cubic-DUCG model is shown in Figure 5, in which the variables in different time layers are treated as different nodes even for a same variable. All variables are state-known evidences indicated in colors, because the cubic-DUCG is generated based on the original DUCG and according to the online received evidences.

The basic idea of the recursive algorithm is to calculate $Pr\{E\}$ layer by layer, which can reduce the computation greatly, given the precondition that all nodes are state-known.

The recursive algorithm can be summarized as follows.

**Definition**: Let $l$ be the layer index, where $l$ is the maximal number of arcs between the *X*-type nodes in layer $l$ and root cause $B_k$ in a DUCG.

For the example in Fig. 4, $X_1$ through $X_4$ are in layer 1 ($l$=1), $X_5$ through $X_8$ are in layer 2 ($l$=2), ···, $X_{17}$ through $X_{19}$ are in layer 5 ($l$=5). In Fig. 5, $X_1$ through $X_5$ and $X_{14}$ are in layer 1 ($l$=1), $X_6$ through $X_8$ are in layer 2 ($l$=2), $X_9$ and $X_{10}$ are in layer 3 ($l$=3), $X_{11}$ and $X_{12}$ are in layer 4 ($l$=4), and $X_{13}$ is in layer 5 (l=5).

Denote the evidence events (state-known *X*-type nodes) in layer $l$ as $E(l)$, we have $E = \bigcap_{l=1}^{l_{max}} E(l)$. When there is no evidence in layer *l*, e.g. layers 1-4 in Figure 4, $E(l)$=1 (complete set). Then we have Eqn.(4).

$$\Pr\{E\} = \sum_j \Pr\{E \mid B_{kj}\}\Pr\{B_{kj}\}$$
$$= \sum_j \Pr\{\bigcap_{l=1}^{l_{max}} E(l) \mid B_{kj}\}\Pr\{B_{kj}\} \quad (4)$$
$$= \sum_j Pr\{E(l_{max}) \mid B_{kj} \bigcap_{l=1}^{l_{max}-1} E(l)\} Pr\{E(l_{max}-1) \mid B_{kj} \bigcap_{l=1}^{l_{max}-2} E(l)\} \cdots$$
$$Pr\{E(2) \mid B_{kj} E(1)\} Pr\{E(1) \mid B_{kj}\}\Pr\{B_{kj}\}$$

In which $Pr\{B_{kj}\} = b_{kj}$ is given during the DUCG construction.

**Assumption 8**[1]: $E(l)$ is independent of $E(l-2)$ through $E(1)$ and $B_{kj}$ conditional on $E(l-1)$.

Under Assumption 8, Eqn.(4) becomes Eqn.(5).

$$\Pr\{E\} = \sum_j Pr\{E(l_{max}) \mid E(l_{max}-1)\} Pr\{E(l_{max}-1) \mid E(l-2)\} \cdots \quad (5)$$
$$Pr\{E(2) \mid E(1)\} Pr\{E(1) \mid B_{kj}\}\Pr\{B_{kj}\}$$

**Assumption 9**: All parents of $E(l)$ are included in $E(l-1)$.

Under Assumption 9, Eqn.(5) becomes Eqn.(6), in which Eqns.(1) and (2) are applied.

$$\Pr\{E\} = \sum_j \Pr\{\bigcap_{X_{nk_n}\in E(l_{max})} X_{nk_n} \mid \bigcap_{X_{ij_i}\in E(l_{max}-1)} X_{ij_i}\} Pr\{\bigcap_{X_{nk_n}\in E(l_{max}-1)} X_{nk_n} \mid \bigcap_{X_{ij_i}\in E(l_{max}-2)} X_{ij_i}\} \cdots$$
$$Pr\{\bigcap_{X_{nk_n}\in E(2)} X_{nk_n} \mid \bigcap_{X_{ij_i}\in E(1)} X_{ij_i}\} Pr\{\bigcap_{X_{nk_n}\in E(1)} X_{nk_n} \mid B_{kj}\}\Pr\{B_{kj}\}$$
$$= \sum_j \Pr\{\bigcap_{X_{nk_n}\in E(l_{max})} \sum_{X_{ij_i}\in E(l_{max}-1)} F_{nk_n;ij_i}\} Pr\{\bigcap_{X_{nk_n}\in E(l_{max}-1)} \sum_{X_{ij_i}\in E(l_{max}-2)} F_{nk_n;ij_i}\} \cdots$$
$$Pr\{\bigcap_{X_{nk_n}\in E(2)} \sum_{X_{ij_i}\in E(1)} F_{nk_n;ij_i}\} Pr\{\sum_{X_{nk_n}\in E(1)} F_{nk_n;ij_i}\}\Pr\{B_{kj}\} \quad (6)$$
$$= \sum_j \left( \prod_{X_{nk_n}\in E(l_{max})} \sum_{X_{ij_i}\in E(l_{max}-1)} f_{nk_n;ij_i} \right)\left( \prod_{X_{nk_n}\in E(l_{max}-1)} \sum_{X_{ij_i}\in E(l_{max}-2)} f_{nk_n;ij_i} \right)\cdots$$
$$\left( \prod_{X_{nk_n}\in E(2)} \sum_{X_{ij_i}\in E(1)} f_{nk_n;ij_i} \right)\left( \prod_{X_{nk_n}\in E(1)} f_{nk_n;kj} \right) b_{kj}$$

In which $f_{nk_n;ij_i} = (r_{n;i}/r_n)a_{nk_n;ij_i}$ and the $\{r\text{-}, a\text{-}\}$-type parameters are given during the DUCG construction.

For a given sub-DUCG$_{kj}$ addressed in Section 2.2.2 in [6], Eqn.(6) becomes Eqn.(7).

---

[1] Assumptions are indexed serially in DUCG papers.

$$\Pr\{E\,|\,\text{sub-DUCG}_{kj}\} = \left(\prod_{X_{nk_n}\in E(l_{\max})}\sum_{X_{ij_i}\in E(l_{\max}-1)} f_{nk_n;ij_i}\right)\left(\prod_{X_{nk_n}\in E(l_{\max}-1)}\sum_{X_{ij_i}\in E(l_{\max}-2)} f_{nk_n;ij_i}\right)\cdots$$
$$\left(\prod_{X_{nk_n}\in E(2)}\sum_{X_{ij_i}\in E(1)} f_{nk_n;ij_i}\right)\left(\prod_{X_{nk_n}\in E(1)} f_{nk_n;kj}\right)b_{kj} \qquad (7)$$

Eqn.(7) is very simple to calculate and frequently used in DUCG-added clinical diagnoses.

In some cases, Assumptions 8 and 9 are not satisfied. For the example in Fig. 5, $X_6$ is the parent of $E(5)=X_{13,j}$ but is not included in $E(4)$. In such cases, we have

$$\Pr\{E\} = \sum_j \Pr\{\bigcap_{X_{nk_n}\in E(l_{\max})} X_{nk_n}\,|\,\bigcap_{V_{ij_i}\in AEB(l_{\max})} V_{ij_i}\}\Pr\{\bigcap_{X_{nk_n}\in E(l_{\max}-1)} X_{nk_n}\,|\,\bigcap_{V_{ij_i}\in AEB(l_{\max}-1)} V_{ij_i}\}\cdots$$
$$\Pr\{\bigcap_{X_{nk_n}\in E(2)} X_{nk_n}\,|\,\bigcap_{V_{ij_i}\in AEB(2)} V_{ij_i}\}\Pr\{\bigcap_{X_{nk_n}\in E(1)} X_{nk_n}\,|\,B_{kj}\}\Pr\{B_{kj}\}$$
$$= \sum_j \Pr\{\bigcap_{X_{nk_n}\in E(l_{\max})V_{ij_i}\in AEB(l_{\max}-1)}\sum F_{nk_n;ij_i}\}\Pr\{\bigcap_{X_{nk_n}\in E(l_{\max}-1)V_{ij_i}\in E(l_{\max}-2)}\sum F_{nk_n;ij_i}\}\cdots$$
$$\Pr\{\bigcap_{X_{nk_n}\in E(2)V_{ij_i}\in E(2)}\sum F_{nk_n;ij_i}\}\Pr\{\bigcap_{X_{nk_n}\in E(1)} F_{nk_n;kj}\}\Pr\{B_{kj}\} \qquad (8)$$
$$= \sum_j \left(\prod_{X_{nk_n}\in E(l_{\max})V_{ij_i}\in AEB(l_{\max})}\sum f_{nk_n;ij_i}\right)\left(\prod_{X_{nk_n}\in E(l_{\max}-1)V_{ij_i}\in AEB(l_{\max})}\sum f_{nk_n;ij_i}\right)\cdots$$
$$\left(\prod_{X_{nk_n}\in E(2)V_{ij_i}\in AEB(2)}\sum f_{nk_n;ij_i}\right)\left(\prod_{X_{nk_n}\in E(1)} f_{nk_n;kj}\right)b_{kj}$$

In which, $AEB(l)$ indicates the directly linked Ancestor Evidence/$B$-type event of the $X$-type evidence node in layer $l$. Eqn.(8) is actually used to calculate the cubic-DUCG presented in [6] and [25]. Figure 5 is an example.

However, in some cases such as Figure 4, Assumptions 8 and 9 are not satisfied and Eqns.(7) and (8) are not applicable. We can only apply Eqn.(4), which involves to expand the evidence nodes to the upper stream layer evidence nodes or $B$-type nodes through intermediate state-unknown nodes such as $X_1$ through $X_{16}$ in Figure 4. As shown in Sec.2.3, the complexity of the expanding is exponential, because of the state combination explosion.

## 2.3. Combination explosion in expanding *E*

It is hard to directly derivate the complexity to expand the evidence nodes in a DUCG. For simplicity, Figure 5 is simplified as Figure 6 to illustrate the complexity, in which only exist one root cause $B_1$ and evidence $X_{17,1}$. This is a fully connected $n\times n$ DUCG with the evidence in the lowest layer.

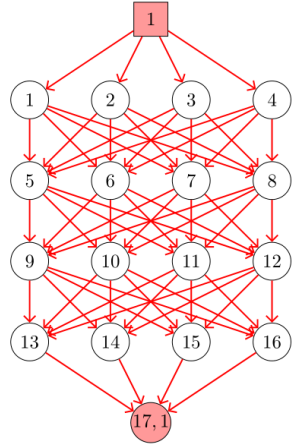

Figure 6: A full joined DUCG with n × n size, where n=5

For each path that expands $X_{17,1}$ to $B_1$, e.g. $X_{17,1} \leftarrow X_{13} \leftarrow X_9 \leftarrow X_5 \leftarrow X_1 \leftarrow B_1$, $n+1$ arcs are referred during calculation, e.g. to calculate $A_{17;13}A_{13;9}A_{9;5}A_{5;1}A_{1;B}$. Suppose that every node has $k$ states, then the dimension of all *A*-type matrices is $k \times k$. Because the time complexity of multiplication between two $k \times k$ matrices is $k^3$, the time complexity for expanding each path is $O(k^3(n+1)) = O(k^3 n)$. In addition, because there are $n^n$ paths to expand, the computation complexity is $O(k^3 n^{n+1})$. This is too large to calculate.

## 2.4. Combination explosion for logic operation

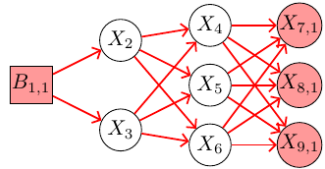

Figure 7: *Example for expression absorb*

In the process of expanding $E(l)$ to layer *l*-1, at first, each evidence $X_{nj}$ in $E(l)$ is expanded separately which is denoted as $Exp_n$, then these $Exp_n$ will be multiplied together. This process of multiplication is exponential, because it needs to handle with the combination of the multiplied $Exp_n$. Meanwhile, as some different evidences in $E(l)$ have same parent variables, some terms in $Exp_n$ can be merged, which is called absorbing. Because the absorbing process needs to check all terms in $Exp_n$, it is also an exponential problem. These two problems cause the combination explosion for logic operation.

Figure 7 shows an example for this combination explosion. $E(3) = X_{7,1}X_{8,1}X_{9,1}$ can be reformed as Eqn.(9) after expanding to layer 2. Each evidence contains 3 terms after expanding, then the full expression contains $3^3 = 27$ terms, which is exponential to the count of nodes in layer 2 and layer 3.

$$
\begin{aligned}
E &= X_{6,1} \cdot X_{7,1} \cdot X_{8,1} \\
&= (F_{7,1;4}X_4 + F_{7,1;5}X_5 + F_{7,1;6}X_6) \cdot (F_{8,1;4}X_4 + F_{8,1;5}X_5 + F_{8,1;6}X_6) \cdot (F_{9,1;4}X_4 + F_{9,1;5}X_5 + F_{9,1;6}X_6) \\
&= F_{9,1;4} \cdot F_{8,1;4} \cdot F_{7,1;4}X_4 + F_{7,1;5} \cdot F_{8,1;5} \cdot F_{9,1;5}X_5 + F_{7,1;6} \cdot F_{8,1;6} \cdot F_{9,1;6}X_6 \\
&+ (F_{7,1;5} \cdot F_{9,1;4} \cdot F_{8,1;4} + F_{8,1;5} \cdot F_{9,1;4} \cdot F_{7,1;4} + F_{9,1;5} \cdot F_{8,1;4} \cdot F_{7,1;4} \\
&+ F_{7,1;5} \cdot F_{8,1;5} \cdot F_{9,1;4} + F_{7,1;5} \cdot F_{9,1;5} \cdot F_{8,1;4} + F_{8,1;5} \cdot F_{9,1;5} \cdot F_{7,1;4})X_4X_5 \\
&+ (F_{7,1;5} \cdot F_{8,1;6} \cdot F_{9,1;6} + F_{7,1;6} \cdot F_{8,1;5} \cdot F_{9,1;6} + F_{7,1;6} \cdot F_{8,1;6} \cdot F_{9,1;5} \\
&+ F_{7,1;5} \cdot F_{8,1;5} \cdot F_{9,1;6} + F_{7,1;5} \cdot F_{8,1;6} \cdot F_{9,1;5} + F_{7,1;6} \cdot F_{8,1;5} \cdot F_{9,1;5})X_5X_6 \\
&+ (F_{7,1;6} \cdot F_{8,1;6} \cdot F_{9,1;4} + F_{7,1;6} \cdot F_{9,1;6} \cdot F_{8,1;4} + F_{8,1;6} \cdot F_{9,1;6} \cdot F_{7,1;4} \\
&+ F_{7,1;6} \cdot F_{9,1;4} \cdot F_{8,1;4} + F_{8,1;6} \cdot F_{9,1;4} \cdot F_{7,1;4} + F_{9,1;6} \cdot F_{8,1;4} \cdot F_{7,1;4})X_4X_6 \\
&+ (F_{7,1;5} \cdot F_{8,1;6} \cdot F_{9,1;4} + F_{7,1;5} \cdot F_{9,1;6} \cdot F_{8,1;4} + F_{7,1;6} \cdot F_{8,1;5} \cdot F_{9,1;4} \\
&+ F_{7,1;6} \cdot F_{9,1;5} \cdot F_{8,1;4} + F_{8,1;5} \cdot F_{9,1;6} \cdot F_{7,1;4} + F_{8,1;6} \cdot F_{9,1;5} \cdot F_{7,1;4})X_4X_5X_6
\end{aligned}
$$
(9)

A new estimation method is proposed for this combination explosion in Sec.3.2. The basic idea is to ignore the terms with high-order small value in the expression, and a new definition called *n*-order *F* is addressed for this algorithm.

## 3. Stochastic simulation algorithm for DUCG

As mentioned in Sec.2, the DUCG inference can be transformed as to calculate $Pr\{E | B_{kj}\}$. Because different root causes are independent with each other, conditional probabilities for them can be calculated separately. Thus, we need only to discuss the calculation for $B_{kj}$.

Because of the expanding of $E$ is too large to calculate, we propose to use sampling to perform the calculation. However, the value of $Pr\{E | B_{2,1}\}$ in DUCG can be too little to sample, e.g. $Pr\{E | B_{2,1}\}$ in figure 8 is $5.626 \times 10^{-14}$, which means too many samples should be taken. To overcome this problem, we present the conditional sampling algorithm as follows, which is based on the recursive algorithm.

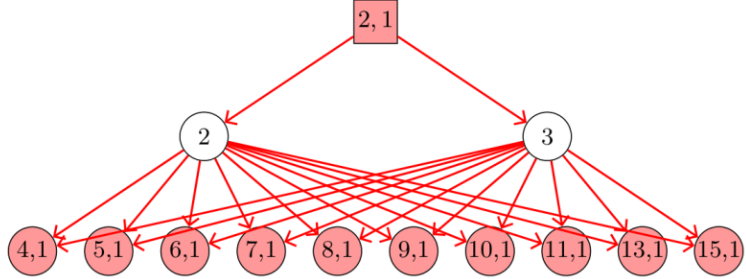

Figure 8: A case of Viral Hepatitis B with 32 evidences

## 3.1. Sampling algorithm for a one evidence case

At first, the case with only one evidence is discussed. The basic sampling algorithm is briefly shown below:

1) Set each state-unknown variable $X_n$ with a random initial state as $X_{n,sn(0)}$;

2) Denote $W_u$ as the set of state-unknown variables;

3) Calculate the conditional distribution of $X_n$ as Eqn.(10), then assign $X_n$ with a new state which is sampled from the distribution as $X_{n,Sn(t)}$, where $S_{n(t)}$ indicates the assigned state of variable $X_n$ in $t_{th}$ sampling cycle and $X_{n,Sn(t)}$ means the event that $X_n = S_{n(t)}$

$$Pr\left\{X_n \mid \bigcap_{X_e \in E} X_e \bigcap_{X_l \in W_u, l \neq n} X_{l,Sl(t-1)}\right\} \qquad (10)$$

4) Calculate the value of $P_t = Pr\{E | \bigcap_{X_n \in W_u} X_n = S_{n(t)}\}$;

5) Get the expectation of $P_t$ as $P(t) = Exp(P_t) = \frac{1}{t}\sum_{j=1}^{t} P_j$;

6) Repeat step 3~5 until the sequence of $P(t)$ convergence;

7) Output the value of $P(t)$ as the sampling result of $Pr\{E | B_{k,j}\}$ and output the value of $N = t$ as the count of sampling cycles.

The key points includes getting the distribution $Pr\left\{X_n \mid \bigcap_{X_e \in E} X_e \bigcap_{X_l \in W_u, l \neq n} X_{l,Sl(t-1)}\right\}$ for updating state-unknown variables' states and getting the value of $P_t = Pr\{E | \bigcap_{X_n \in W_u} X_{n,Sn(t)}\}$.

Because all state-unknown variables have been assigned states during previous sampling cycle, the expanding process for $X_n$ will end at its parent variables according to the recursive algorithm. Denote that $W_{pn}$ is the set of parents of $X_n$, then the conditional distribution of $X_n$ can be sampled with Eqn.(11) .

$$Pr\left\{X_n \mid \bigcap_{X_e \in E} X_e \bigcap_{X_l \in W_u, l \neq n} X_{l, Sl(t-1)}\right\}$$

$$= Pr\left\{X_n^t = S \mid \bigcap_{V_i \in W_{pn}} V_i = s_{i(t-1)}\right\}$$

$$= Pr\left\{\sum_{V_i \in W_p} F_{ns;i,Si(t-1)} V_{i,Si(t-1)} \mid \bigcap_{V_i \in W_{pn}} V_{i,Si(t-1)}\right\} \quad (11)$$

$$= \sum_i \frac{r_{n;i}}{r_n} A_{ns;,iSi(t-1)}$$

Similarly, denote $W_{pE}$ as the set of parents variables of $E$, we have:

$$\begin{aligned} P_t &= Pr\{E \mid \bigcap_{X_n \in W_u} X_{n,Sn(t)}\} \\ &= Pr\{E \mid \bigcap_{V_i \in W_{pE}} V_{i,Si(t)}\} \end{aligned} \quad (12)$$

Example in Figure 9 is presented to verify the algorithm. Denote that $V_1$ is $X_i$'s parents, $X_2, X_3...X_i$ are the parents of evidence $E_{4,1}$. As shown in Algorithm 1:

$$P_t = Pr\{E_{4,1} \mid X_{2,S2(t)} X_{3,S3(t)}...\}$$

Then, we have:

$$P = Exp(P_t) = \frac{1}{N} \sum_{t=1}^{N} Pr\{E_{4,1} \mid X_{2,S2(t)} X_{3,S3(t)}...\}$$

Because $X_2^t, X_3^t$ are sampled conditioned on $V$, the frequencies of event $X_i = S$ will follow the probabilities $1/N \cdot freq(X_i = S) = Pr\{X_{is} \mid V_1\}$, so $P$ can be reformed as Eqn.(13).

$$\begin{aligned} P &= \sum_{X_2=i} \sum_{X_3=j} Pr\{E_{4,1} \mid X_{2,i} X_{3,j}\} \frac{1}{N} freq(X_{2,i} X_{3,j}) \\ &= \sum_{X_2=i} \sum_{X_3=j} Pr\{E_{4,1} \mid X_{2,i} X_{3,j}\} Pr\{X_{2,i} X_{3,j} \mid V_1\} \\ &= \sum_{X_2=i} \sum_{X_3=j} Pr\{E_{4,1} \mid X_{2,i} X_{3,j}\} Pr\{X_{2,i} \mid V_1\} Pr\{X_{3,j} \mid V_1\} \\ &= \sum_{X_2=i} \sum_{X_3=j} (F_{4,1;2,i} + F_{4,1;3,j}) F_{2,i;v_1} F_{3,j;v_1} \\ &= \sum_{X_2=i} \sum_{X_3=j} (F_{4,1;2,i} F_{2,i;v_1} F_{3,j;v_1} + F_{4,1;3,j} F_{2,i;v_1} F_{3,j;v_1}) \\ &= \sum_{X_2=i} F_{4,1;2,i} F_{2,i;v_1} \sum_{X_3=j} F_{3,j;v_1} \\ &+ \sum_{X_3=j} F_{4,1;3,j} F_{3,j;v_1} \sum_{X_2=i} F_{2,i;v_1} \end{aligned} \quad (13)$$

Meanwhile, the inference result of traditional algorithm is Eqn.(14), which is the same as Eqn. (13).

$$\begin{aligned} Pr\{E_{4,1} \mid V_1\} &= Pr\{(F_{4,1;2} F_{2;1} + F_{4,1;3} F_{3;1}) V_1 \mid V_1\} \\ &= \sum_{X_2=i} F_{4,1;2,i} F_{2,i;v_1} \sum_{X_3=j} F_{3,j;v_1} \\ &+ \sum_{X_3=j} F_{4,1;3,j} F_{3,j;v_1} \sum_{X_2=i} F_{2,i;v_1} \end{aligned} \quad (14)$$

$E_{4,1}$ is state-known, so the expanding of its children will only refer to its known state 1 according to the recursive algorithm. Therefore, evidences variables will not require states updating during sampling cycles.

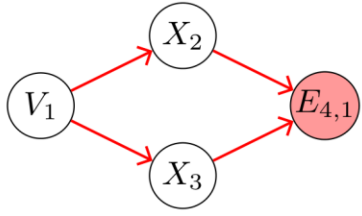

Figure 9: Basic example for the sampling algorithm

If $E_{4,1}$ is replaced by a state-unknown variable $X_l$, frequency of $X_l = S$ during cycles will also meet $1/N \cdot freq(X_l = S) = Pr\{X_{l,S} | V_1\}$ according to Eqn.(13). It can be figured out that frequency of its child's sampling states also satisfies this equation with similar derivation.

In conclusion, frequency of each single node $X_n$'s sampling states satisfies Eqn.(15).

$$\frac{1}{N} freq(X_n = S) = Pr\{X_{n,i} | V_1\} \tag{15}$$

## 3.2. Multi evidences situation and Absorbing problem

In this section, the graph containing multi evidences is discussed. Because different evidences can refer to same parents, the absorbing process is required for expanding expression. So, multiplying the result in Sec.3.1 for each $X_{n,s}$ in $E$ together will lead to systemic errors.

The example in Figure 7 is used to show this systemic error. The result will be Eqn.(16) if the evidences are sampled separately with Algorithm 1 and then multiplied together.

$$\bigcap_{X_{n,k} \in E} Pr\{X_{n,k} | \bigcap_{X_e \in E} X_e \bigcap_{X_l \in W_u, l \neq n} X_{l,Sl(t-1)}\}$$
$$= Pr\{X_{7,1} | \bigcap_{V_i \in W_{p7}} V_i = s_{i(t-1)}\} Pr\{X_{8,1} | \bigcap_{V_i \in W_{p8}} V_i = s_{i(t-1)}\} Pr\{X_{9,1} | \bigcap_{V_i \in W_{p9}} V_i = s_{i(t-1)}\} \tag{16}$$

The equation in Eqn. (16) can be reformed as Eqn.(17) according to Eqn.(15).

$$\begin{aligned}
&\frac{1}{N}\sum_{t=1}^{N}(\bigcap_{X_{n,k} \in E} Pr\{X_{n,k} | \bigcap_{X_e \in E} X_e \bigcap_{X_l \in W_u, l \neq n} X_{l,Sl(t-1)}\})\\
&\quad (\sum_{s_4} F_{7,1;4,s_4} Pr\{X_{4,s_4} | B_{1,1}\} + \sum_{s_5} F_{7,1;5,s_5} Pr\{X_{5,s_5} | B_{1,1}\} + \sum_{s_6} F_{7,1;6,s_6} Pr\{X_{6,s_6} | B_{1,1}\})\\
&= \cdot(\sum_{s_4} F_{8,1;4,s_4} Pr\{X_{4,s_4} | B_{1,1}\} + \sum_{s_5} F_{8,1;5,s_5} Pr\{X_{5,s_5} | B_{1,1}\} + \sum_{s_6} F_{8,1;6,s_6} Pr\{X_{6,s_6} | B_{1,1}\})\\
&\quad \cdot(\sum_{s_4} F_{9,1;4,s_4} Pr\{X_{4,s_4} | B_{1,1}\} + \sum_{s_5} F_{9,1;5,s_5} Pr\{X_{5,s_5} | B_{1,1}\} + \sum_{s_6} F_{9,1;6,s_6} Pr\{X_{6,s_6} | B_{1,1}\})\\
&\quad (Pr\{F_{7,1;4}X_4 | B_{1,1}\} + Pr\{F_{7,1;5}X_5 | B_{1,1}\} + Pr\{F_{7,1;6}X_6 | B_{1,1}\})\\
&= \cdot(Pr\{F_{8,1;4}X_4 | B_{1,1}\} + Pr\{F_{8,1;5}X_5 | B_{1,1}\} + Pr\{F_{8,1;6}X_6 | B_{1,1}\})\\
&\quad \cdot(Pr\{F_{9,1;4}X_4 | B_{1,1}\} + Pr\{F_{9,1;5}X_5 | B_{1,1}\} + Pr\{F_{9,1;6}X_6 | B_{1,1}\})
\end{aligned} \tag{17}$$

In traditional recursive algorithm, $E$ is expanded as Eqn.(18), then $Pr\{E | B_{1,1}\}$ is calculated as Eqn.(19). The difference between Eqn. (17) and Eqn. (19) shows the influence of absorbing process. Therefore, some adjustment is needed for the basic sampling algorithm.

$$\begin{aligned}
E &= X_{7,1} \cdot X_{8,1} \cdot X_{9,1} \\
&= \begin{array}{l}
F_{9,1;4} \cdot F_{8,1;4} \cdot F_{7,1;4} X_4 + F_{7,1;5} \cdot F_{8,1;5} \cdot F_{9,1;5} X_5 + F_{7,1;6} \cdot F_{8,1;6} \cdot F_{9,1;6} X_6 \\
+ (F_{7,1;5} \cdot F_{9,1;4} \cdot F_{8,1;4} + F_{8,1;5} \cdot F_{9,1;4} \cdot F_{7,1;4} + F_{9,1;5} \cdot F_{8,1;4} \cdot F_{7,1;4} \\
+ \cdot F_{7,1;5} \cdot F_{8,1;5} \cdot F_{9,1;4} + F_{7,1;5} \cdot F_{9,1;5} \cdot F_{8,1;4} + F_{8,1;5} \cdot F_{9,1;5} \cdot F_{7,1;4}) X_4 X_5 \\
+ (F_{7,1;5} \cdot F_{8,1;6} \cdot F_{9,1;6} + F_{7,1;6} \cdot F_{8,1;5} \cdot F_{9,1;6} + F_{7,1;6} \cdot F_{8,1;6} \cdot F_{9,1;5} \\
+ F_{7,1;5} \cdot F_{8,1;5} \cdot F_{9,1;6} + F_{7,1;5} \cdot F_{8,1;6} \cdot F_{9,1;5} + F_{7,1;6} \cdot F_{8,1;5} \cdot F_{9,1;5}) X_5 X_6 \\
+ (F_{7,1;6} \cdot F_{8,1;6} \cdot F_{9,1;4} + F_{7,1;6} \cdot F_{9,1;6} \cdot F_{8,1;4} + F_{8,1;6} \cdot F_{9,1;6} \cdot F_{7,1;4} \\
+ F_{7,1;6} \cdot F_{9,1;4} \cdot F_{8,1;4} + F_{8,1;6} \cdot F_{9,1;4} \cdot F_{7,1;4} + F_{9,1;6} \cdot F_{8,1;4} \cdot F_{7,1;4}) X_4 X_6 \\
+ (F_{7,1;5} \cdot F_{8,1;6} \cdot F_{9,1;4} + F_{7,1;5} \cdot F_{9,1;6} \cdot F_{8,1;4} + F_{7,1;6} \cdot F_{8,1;5} \cdot F_{9,1;4} \\
+ F_{7,1;6} \cdot F_{9,1;5} \cdot F_{8,1;4} + F_{8,1;5} \cdot F_{9,1;6} \cdot F_{7,1;4} + F_{8,1;6} \cdot F_{9,1;5} \cdot F_{7,1;4}) X_4 X_5 X_6
\end{array}
\end{aligned} \quad (18)$$

$$\begin{aligned}
&Pr\{E \mid B_{1,1}\} \\
&= \begin{array}{l}
Pr\{F_{9,1;4} \cdot F_{8,1;4} \cdot F_{7,1;4} X_4 \mid B_{1,1}\} + Pr\{F_{7,1;5} \cdot F_{8,1;5} \cdot F_{9,1;5} X_5 \mid B_{1,1}\} \\
+ Pr\{F_{7,1;6} \cdot F_{8,1;6} \cdot F_{9,1;6} X_6 \mid B_{1,1}\} + Pr\{(F_{7,1;5} \cdot F_{9,1;4} \cdot F_{8,1;4} \\
+ F_{8,1;5} \cdot F_{9,1;4} \cdot F_{7,1;4} + F_{9,1;5} \cdot F_{8,1;4} \cdot F_{7,1;4} + \cdot F_{7,1;5} \cdot F_{8,1;5} \cdot F_{9,1;4} \\
+ F_{7,1;5} \cdot F_{9,1;5} \cdot F_{8,1;4} + F_{8,1;5} \cdot F_{9,1;5} \cdot F_{7,1;4}) X_4 X_5 \mid B_{1,1}\} \\
+ Pr\{(F_{7,1;5} \cdot F_{8,1;6} \cdot F_{9,1;6} + F_{7,1;6} \cdot F_{8,1;5} \cdot F_{9,1;6} + F_{7,1;6} \cdot F_{8,1;6} \cdot F_{9,1;5} \\
+ F_{7,1;5} \cdot F_{8,1;5} \cdot F_{9,1;6} + F_{7,1;5} \cdot F_{8,1;6} \cdot F_{9,1;5} + F_{7,1;6} \cdot F_{8,1;5} \cdot F_{9,1;5}) X_5 X_6 \mid B_{1,1}\} \\
+ Pr\{(F_{7,1;6} \cdot F_{8,1;6} \cdot F_{9,1;4} + F_{7,1;6} \cdot F_{9,1;6} \cdot F_{8,1;4} + F_{8,1;6} \cdot F_{9,1;6} \cdot F_{7,1;4} \\
+ F_{7,1;6} \cdot F_{9,1;4} \cdot F_{8,1;4} + F_{8,1;6} \cdot F_{9,1;4} \cdot F_{7,1;4} + F_{9,1;6} \cdot F_{8,1;4} \cdot F_{7,1;4}) X_4 X_6 \mid B_{1,1}\} \\
+ Pr\{(F_{7,1;5} \cdot F_{8,1;6} \cdot F_{9,1;4} + F_{7,1;5} \cdot F_{9,1;6} \cdot F_{8,1;4} + F_{7,1;6} \cdot F_{8,1;5} \cdot F_{9,1;4} \\
+ F_{7,1;6} \cdot F_{9,1;5} \cdot F_{8,1;4} + F_{8,1;5} \cdot F_{9,1;6} \cdot F_{7,1;4} + F_{8,1;6} \cdot F_{9,1;5} \cdot F_{7,1;4}) X_4 X_5 X_6 \mid B_{1,1}\}
\end{array}
\end{aligned} \quad (19)$$

An initiative way is to expanded evidences $E$ to the root cause $B_{kj}$, and then sample on it directly. However, this method cannot take the advantage of conditional sampling algorithm in Sec.3.1 and is low efficient.

Faced with the problem, a new cut-off estimation algorithm is proposed.

Because the first three terms in Eqn.(18) contains only one $X_n$, they satisfy Eqn. (15), so they can be replaced by events $X_4^t$, $X_5^t$ and $X_6^t$ during calculation in Eqn. (19). Then these terms can be substituted by numerical values. Meanwhile, the following terms require more processing. The forth term containing $X_4X_5$ is used for an example. It can be expanded to Eqn.(20).

$$\begin{aligned}
C_{45} \cdot X_4 X_5 &= C_{45}[F_{4;2} F_{5;2} X_2 + F_{4;3} F_{5;3} X_3 \\
&\quad + (F_{4;2} F_{5;3} + F_{4;3} F_{5;2}) X_2 X_3] \\
&= C_{45}[F_{4;2} F_{5;2} X_2 + F_{4;3} F_{5;3} X_3 \\
&\quad + (F_{4;2} F_{5;3} + F_{4;3} F_{5;2}) F_{2;1,1} F_{3;1,1} B_{1,1}])
\end{aligned} \quad (20)$$

For simplicity, $C_{45}$ is on behalf of the sum-of-products of $F$-type in this term:

$$C_{45} = \begin{array}{l} F_{7,1;5} \cdot F_{9,1;4} \cdot F_{8,1;4} + F_{8,1;5} \cdot F_{9,1;4} \cdot F_{7,1;4} \\ + F_{9,1;5} \cdot F_{8,1;4} \cdot F_{7,1;4} + \cdot F_{7,1;5} \cdot F_{8,1;5} \cdot F_{9,1;4} \\ + F_{7,1;5} \cdot F_{9,1;5} \cdot F_{8,1;4} + F_{8,1;5} \cdot F_{9,1;5} \cdot F_{7,1;4} \end{array}$$

Similarly, terms containing only $X_2$ or $X_3$ in Eqn.(20) can be replaced by the event $X_2^t$ and $X_3^t$ in $t_{th}$ sampling cycle. The third term containing $X_2X_3$ is similar to the term containing $X_4X_5$, so it needs to be expanded further as $(F_{4;2} F_{5;3} + F_{4;3} F_{5;2}) F_{2;1,1} F_{3;1,1} B_{1,1}$. Then its numerical value can be calculated.

As all terms in expanding expression have the form $\prod F \prod X$, the definition of n-order F is given.

**Definition:** For a term $\prod F \prod X$ in the expanding expression, if the count of F-type events in products $\prod F$ is $n$, this term is $n$-order F.

The expanding process in Eqn.(20) ends after two steps, but more complex terms will needs more expanding steps. The term with $X_4X_5X_6$ in Eqn. (18) is presented as an example, and the expanding expression is Eqn.(21).

$$\begin{aligned} & C_{456} \cdot X_4 X_5 X_6 \\ = & \begin{array}{l} C_{456} \cdot [F_{4;2}F_{5;2}F_{6;2}X_2 + F_{4;3}F_{5;3}F_{6;3}X_3 \\ +(F_{4;2}F_{5;3}F_{6;3} + F_{4;3}F_{5;2}F_{6;3} + F_{4;3}F_{5;3}F_{6;2} \\ +F_{4;2}F_{5;2}F_{6;3} + F_{4;2}F_{5;3}F_{6;2} + F_{4;3}F_{5;2}F_{6;2})X_2X_3] \end{array} \\ = & \begin{array}{l} C_{456} \cdot [F_{4;2}F_{5;2}F_{6;2}X_2 + F_{4;3}F_{5;3}F_{6;3}X_3 \\ +(F_{4;2}F_{5;3}F_{6;3} + F_{4;3}F_{5;2}F_{6;3} + F_{4;3}F_{5;3}F_{6;2} \\ +F_{4;2}F_{5;2}F_{6;3} + F_{4;2}F_{5;3}F_{6;2} + F_{4;3}F_{5;2}F_{6;2})F_{2;1,1}F_{3;1,1}B_{1,1}] \end{array} \end{aligned} \quad (21)$$

Similarly, $C_{456}$ represents for:

$$C_{456} = \begin{array}{l} F_{7,1;5} \cdot F_{8,1;6} \cdot F_{9,1;4} + F_{7,1;5} \cdot F_{9,1;6} \cdot F_{8,1;4} \\ +F_{7,1;6} \cdot F_{8,1;5} \cdot F_{9,1;4} + F_{7,1;6} \cdot F_{9,1;5} \cdot F_{8,1;4} \\ +F_{8,1;5} \cdot F_{9,1;6} \cdot F_{7,1;4} + F_{8,1;6} \cdot F_{9,1;5} \cdot F_{7,1;4} \end{array}$$

The term in Eqn.(20) with only $X_2$ is $C_{45}F_{4;2}F_{5;2}X_2$, while the similar one in Eqn.(21) is $C_{456}F_{4;2}F_{5;2}F_{6;2}X_2$. Both of $C_{45}$ and $C_{456}$ are sums of three 3-order F terms, i.e., $F_iF_jF_k$. Then, $C_{45}F_{4;2}F_{5;2}X_2$ is 5-order F and $C_{456}F_{4;2}F_{5;2}F_{6;2}X_2$ is 6-order F. Because Eqn. (20) is expanded from $C_{45} \cdot X_4X_5$ and Eqn.(21) is expanded from $C_{456} \cdot X_4X_5X_6$, the more X variables a term contains, the higher order F it will be after the next step expanding.

Secondly, the term $C_{45}F_{4;2}F_{5;2}X_2$ is 5-order F, while the term $C_{45}(F_{4;2}F_{5;3} + F_{4;3}F_{5;2})X_2X_3$ will becomes 7-order F as $C_{45}(F_{4;2}F_{5;3} + F_{4;3}F_{5;2})F_{2;1,1}F_{3;1,1}B_{1,1}$ after next step expanding. Therefore, the more layer a term has been expanded, the higher order F it is.

Multiply for F events follows the AND/multiplication matrix operator specially defined in Corollary 15 in [2], in brief, elements in the result matrix can be represents as sums-of-products of $a$ parameters' product, i.e. $\sum \prod_i^n a_i$, where $n$ is the count of matrices.

As $a_i$ parameter are elements in the relation matrices, they are all less than one. Therefore, the higher n-order F a term is, the less value it is. As a result, terms with too large n-order F can be ignored due to their high order small values.

Two hyper parameters are denoted for this estimation. $IG_x$ defines the threshold for count of $X-$ type variable, and $IG_{layer}$ defines the threshold for count of layers to expand from evidences. All terms beyond the threshold will ignored during calculation. These two thresholds are intuitive and simple, and are verified by the examples in Sec.4.3.

This algorithm is not perfect because the choosing of two hyper parameters is difficult. Better methods will be addressed in future works.

### 3.3. Time Complexity of Sampling Algorithm

In this section, an analysis on time complexity is proposed based on two examples to show the advantage of new algorithm over the traditional one.

### 3.3.1. Stopping rule for sampling

$\epsilon - \delta$ estimate is a common tool for the evaluation of sampling algorithm, the sampling cycles stops after meet the condition shown below.

$$P[\phi(1+\epsilon)^{-1} < \mu < \phi(1+\epsilon)] \geq 1-\Delta$$

$\mu$ is the sampling result and $\phi$ is the exact result of the problem, $\epsilon$ is the error ratio expected for the algorithm, $1-\Delta$ is confidence coefficient and $\epsilon, \Delta < 1$.

Several researches have been proposed for sampling algorithm for BN.

In [18], zero-one estimate is applied and the cycles required is:

$$N = \frac{4}{\phi \epsilon^2} \ln \frac{2}{\Delta}$$

In [16], the sampling process is modeled as Bernoulli process, then the cycles required is:

$$N = \frac{7}{\epsilon^2 \mu} \ln \frac{4}{\Delta}$$

These analyses can be applied to DUCG with few changes. However, $\mu$ or $\phi$ is in denominator for both of these estimations, as $\mu(\phi)$ could reach $10^{-15}$ in DUCG, $N$ will be too large with these constraint conditions. So, these stopping rule is not practical for DUCG.

As mentioned in [19], it's difficult to determine the convergence of MCMC algorithm theoretically. Therefore, a method based on stochastic method is proposed in this section. According to the law of large numbers and central-limit theorem, $\mu$ fits the normal distribution i.e. $\mu \sim N\left(\mu, \frac{\delta^2}{n}\right)$ after sufficient sampling cycles, where $n$ indexes the count of cycles.

Denote the burn-in value as $b$ which means ignoring first $b$ cycles, the window-width value $\omega$ which means calculating the mean value and variance of last $\omega$ cycles. Then the halting problem algorithm is:

1) Set an upper limit of the sampling cycle as $Cycle_{max}$ in case that the sampling cannot convergence;
2) Execute the basic sampling algorithm for $b+\omega$ cycles;
3) Execute the sampling algorithm for one cycle, record the current cycle as the $t_{th}$ cycle;
4) Calculate the means of $P_t$ in last $\omega$ sampling cycles as:

$$\mu_t = \frac{1}{\omega} \sum_{j=t-\omega}^{t} p_j$$

5) Calculate the standard deviation of $P_t$ in last $\omega$ sampling cycles as:

$$\delta_t = [\frac{1}{\omega} \sum_{j=t-\omega}^{t} (p_j - \mu)^2]^{\frac{1}{2}}$$

6) Repeat step 3~5 until $\mu_t$ and $\delta_t$ satisfy:

$$2\int_{(1+\epsilon)\mu_t}^{\infty} N(\mu_t, \delta_t^2) dx < \Delta$$

7) Output the value of $\mu_t$ as sampling result and $N = t$

Then, the sampling algorithm takes around $N > b + \omega + \left(\frac{c\delta}{\epsilon\mu}\right)^2$ cycles. $c = Q^{-1}(\Delta/2)$, where $Q$ is the right tail function of normal distribution; a brief relation of $c$ and $\Delta$ value is presented in Table 4.

| Δ | 31.7% | 4.6% | 0.3% |
|---|-------|------|------|
| c | 1     | 2    | 3    |

Table 4: *Value of Δ and c*

Because $b+\omega \ll N$, and $\delta, \mu$ have similar orders of magnitude in most cases, $N > \left(\frac{c\delta}{\epsilon\mu}\right)^2$ can be used as a proper halting rule for the algorithm, and value of $N$ is not sensitive to either scale of graph or value of $\phi$. Because the diagnoses focus on the sort of root causes' probability instead of the exact numerical value, about 5% error ratio will be enough for practical use.

Because the law of large numbers will lead to some errors when $N$ is not large enough, the practical errors of algorithm can be larger than $\epsilon$ in some cases. As parameter $\epsilon$ and the real error ration of sampling process is positive correlation, using a smaller $\epsilon$ can be helpful.

Then we can get the performance of the algorithm by multiplying the computation complexity during each cycle and $N$ together.

### 3.3.2. Performance for expanding process

This section focuses on the performance of expanding process, and the $n \times n$ full-joined model in Figure 6 is used for a derivation.

As it is shown in Sec.3.1, the expanding process required for updating a single node $X_i$ will only refer to its parents. Therefore, this scheme requires $kn$ steps calculation, where $k$ is count of $X_i$'s states. Because all of $n^2$ state-unknown variables need to be updated, the time complexity is $O(kn^3)$ for the updating process.

In conclusion, the sampling algorithm requires $N > \left(\frac{c\delta}{\epsilon\mu}\right)^2$ loops and needs $O(kn^3)$ calculations in each loop, so the time complexity of the whole algorithm is Eqn.(22)

$$O\left(kn^3 \left(\frac{c\delta}{\epsilon\mu}\right)^2\right) \quad (22)$$

Meanwhile, time complexity of traditional method is $O(k^3 n^{n+1})$ as shown in Sec.2.3. Therefore, the sampling algorithm manages to reduce the complex of expanding process from exponential to polynomial.

### 3.3.3. Performance for logic operation

In the absorbing process, the new algorithm can simplify the expression by applying the estimate method proposed in Sec.3.2. In comparison, the traditional algorithm needs to handle with the full-length expression.

The example in Figure 10 which has $n$ layers with 3n $X$-nodes in each layer is presented to show this advantage.

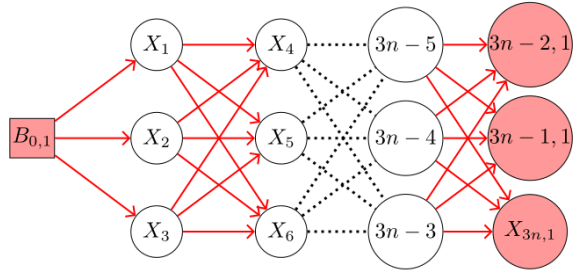

Figure 10: *N layers model with 3 nodes in each layer*

The expanding expression for $E = X_{3n-2,1} X_{3n-1,1} X_{3n,1}$ is similar to Eqn.(9), it has 27 terms, which includes 3 one-$X$ terms, 18 two-$X$ terms and 6 three-$X$ terms. One-$X$ term means this term contains only one $X$-type variable, e.g., $F_{3n-5;3n-8} X_{3n-8}$.

In the traditional algorithm, the whole expression needs to be expanded to the above layers. In one step of expanding, a single one-$X$ term will be expanded to 3 one-x terms, for example:

$$X_{3n-5} = F_{3n-5;3n-8}X_{3n-8} + F_{3n-5;3n-7}X_{3n-7} + F_{3n-5;3n-6}X_{3n-6}$$

The two $X$ terms will be reformed as an expression with 9 terms after one step expanding, and it consists of 3 one $X$ terms and 6 two $X$ terms, e.g.,

$$\begin{aligned}
&X_{3n-3}X_{3n-4} \\
&= (F_{3n-3;3n-8}X_{3n-8} + F_{3n-3;3n-7}X_{3n-7} + F_{3n-3;3n-6}X_{3n-6}) \\
&\quad \cdot (F_{3n-4;3n-8}X_{3n-8} + F_{3n-4;3n-7}X_{3n-7} + F_{3n-4;3n-6}X_{3n-6}) \\
&= F_{3n-3;3n-6}F_{3n-4;3n-6}X_{3n-6} + F_{3n-3;3n-7}F_{3n-4;3n-7}X_{3n-7} + F_{3n-3;3n-8}F_{3n-4;3n-8}X_{3n-8} \\
&\quad + (F_{3n-3;3n-6}F_{3n-4;3n-7} + F_{3n-3;3n-7}F_{3n-4;3n-6})X_{3n-6}X_{3n-7} + (F_{3n-3;3n-6}F_{3n-4;3n-8} \\
&\quad + F_{3n-3;3n-8}F_{3n-4;3n-6})X_{3n-6}X_{3n-8} + (F_{3n-3;3n-7}F_{3n-4;3n-8} + F_{3n-3;3n-8}F_{3n-4;3n-7})X_{3n-7}X_{3n-8}
\end{aligned} \quad (23)$$

The expanding expression for three terms is similar to Eqn. (9), it has 27 terms, and 3 of them are one-X terms, 18 terms are two-X term and 6 terms are three-X terms.

For convenience, we denote the count of one-$X$ terms in $i_{th}$ step expanding expression as $a_i$, count of two-$X$ terms as $b_i$ and three-$X$ terms as $c_i$. So $a_0 = 0, b_0 = 0, c_0 = 1$, and $a_{i+1} = 3(a_i + b_i + c_i)$, $b_{i+1} = 6b_i + 18c_i$, $c_{i+1} = 6c_i$. With these conditions, it can be figured that:

$$c_n = 6^n \quad (24)$$

$$\begin{aligned}
b_n &= 6b_{n-1} + 18c_{n-1} \\
&= 6b_{n-1} + 3c_n \\
&= 3(n-1)6^n
\end{aligned} \quad (25)$$

$$\begin{aligned}
a_n &= 3(a_{n-1} + b_{n-1} + c_{n-1}) \\
&= 3^{n-1}a_1 + \sum_{i=1}^{n-1} 3^{n-i}(3n-2)2^{i-1}3^{i-1} \\
&= (3n-2)6^n - 7 \times 3^n
\end{aligned} \quad (26)$$

Therefore, the expression will have $a_n + b_n + c_n = 2(3n-2)6^n - 7 \times 3^n$ terms when $E$ is expanded to $X_1X_2X_3$.

The terms with largest $F-order$ are the expanding result of three $x$ terms in every layer, and they contain $3n$ $F$ events. Meanwhile, the one-$X$ term will be $n-order$ $F$. Therefore, the time complexity for calculating the value of each term is $O(n)$. So, the time complexity of the traditional algorithm is:

$$O(n(3n-2)6^n - 7n \times 3^n)$$

For simplify:

$$O(n^2 e^n).$$

The situation is simpler for the sampling algorithm with cutting off algorithm in Sec.3.2. As all one-$X$ terms can be substituted by numerical value in each step expanding, the number of one $X$ terms changed to $a'_i = 3(b_{i-1} + c_{i-1})$ in $i_{th}$ expanding step, while $b_i, c_i$ stay the same as the traditional method. Because the expanding stops after $IG_{layer}$ steps, count of terms in sampling algorithm is:

$$\begin{aligned}
&b_{IG_{layer}} + c_{IG_{layer}} + \sum_{i=1}^{IG_{layer}-1} 3(b_i + c_i) \\
&= (3IG_{layer} - 2)6^{IG_{layer}} + 3\sum_{i=1}^{IG_{layer}-1}(3i-2)6^i \\
&= (4.2IG_{layer} - 3.44)6^{IG_{layer}} + 0.24
\end{aligned} \quad (19)$$

Like the traditional algorithm, terms are $3IG_{layer}$-order $F$ or $2IG_{layer}$-order $F$ after the expanding, the time complexity is:

$$O(IG_{layer}(4.2IG_{layer}-3.44)6^{IG_{layer}}+0.24IG_{layer})$$

For simplify:

$$O(IG_{layer}^2 e^{IG_{layer}})$$

Although the complexity of sampling algorithm is still exponential, $IG_{layer}$ is less than $n$ so the time consumption will be less. For instance, if $IG_{layer}=1/2n$ and $n\geq 6$, the traditional expanding will cost 1500 times time than expanding process during one sampling cycle. Therefore, a sampling process ending within 1500 cycles will be faster than the transitional method. The effects of $IG_{len}$ is similar to $IG_{layer}$.

The example shown in Sec.4.3 runs 3 times faster than the traditional algorithm with the boosts of both $IG_{layer}$ and $IG_{len}$. It provides a practical verification that the estimate scheme helps to increase the efficiency of sampling in both expanding and logic operation.

# 4. Application of DUCG sampling algorithm

Three examples for DUCG model are presented in this section. The first one is the multi evidences model illustrated in Sec.2.4, which provide a verification for the estimate method of sampling method proposed in Sec.3.2. The second one is a set of n-size full-joined model like Figure 6 with randomly assigned parameters, and it shows the accuracy and performance for the basic sampling algorithm shown in Sec.3.1. The last one is adapted from a model for Viral Hepatitis B deployed in hospitals, and it shows the performance for the new algorithm in real application. Note that all examples have been simplified and contain only one root cause.

## 4.1. A compact example

Figure 11 is a simple DUCG with only one root cause $B_{1,1}$ and 9 $X$-type variables in three layers. As it is shown, $E=X_{7,1}X_{8,1}X_{9,1}$.

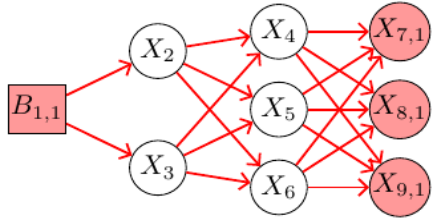

Figure 11:A compact DUCG

For simplify, causal relations $r_{ij}$ for all of $F$-type events are assigned 1, while the relation matrices $A_{ij}$ are assigned as following, they are generated randomly and then normalized:

$$r_{2,1}=1.00, r_{3,1}=1.00$$

$$A_{2;1}=\begin{pmatrix}0.1890 & 0.2490\\0.3440 & 0.4200\\0.4670 & 0.3310\end{pmatrix}, A_{3;1}=\begin{pmatrix}0.7850 & 0.6390\\0.2150 & 0.3610\end{pmatrix}$$

$$r_{4,2}=1.00, r_{4,3}=1.00$$

$$A_{4;2}=\begin{pmatrix}0.9080 & 0.7730 & 0.4440\\0.0920 & 0.2270 & 0.5560\end{pmatrix}, A_{4;3}=\begin{pmatrix}0.5970 & 0.1770\\0.4030 & 0.8230\end{pmatrix}$$

$$r_{5,2}=1.00, r_{5,3}=1.00$$

$$A_{5;2}=\begin{pmatrix}0.1810 & 0.2030 & 0.5180\\0.8190 & 0.7970 & 0.4820\end{pmatrix}, A_{5;3}=\begin{pmatrix}0.0910 & 0.2110\\0.9090 & 0.7890\end{pmatrix}$$

$$r_{6,2} = 1.00, r_{6,3} = 1.00$$

$$A_{6;2} = \begin{pmatrix} 0.5640 & 0.2390 & 0.5600 \\ 0.4360 & 0.7610 & 0.4400 \end{pmatrix}, A_{6;3} = \begin{pmatrix} 0.4760 & 0.6420 \\ 0.5240 & 0.3580 \end{pmatrix}$$

$$r_{7,4} = 1.00, r_{7,5} = 1.00, r_{7,6} = 1.00$$

$$A_{7;4} = \begin{pmatrix} 0.0100 & 0.3030 \\ 0.9900 & 0.6970 \end{pmatrix}, A_{7;5} = \begin{pmatrix} 0.4660 & 0.9520 \\ 0.5340 & 0.0480 \end{pmatrix}, A_{7;6} = \begin{pmatrix} 0.4990 & 0.7070 \\ 0.5010 & 0.2930 \end{pmatrix}$$

$$r_{8,4} = 1.00,, r_{8,5} = 1.00, r_{8,6} = 1.00$$

$$A_{8;4} = \begin{pmatrix} 0.5170 & 0.4750 \\ 0.4830 & 0.5250 \end{pmatrix}, A_{8;5} = \begin{pmatrix} 0.7490 & 0.1190 \\ 0.2510 & 0.8810 \end{pmatrix}, A_{8;6} = \begin{pmatrix} 0.5020 & 0.5100 \\ 0.4980 & 0.4900 \end{pmatrix}$$

$$r_{9,4} = 1.00, r_{9,5} = 1.00 \, r_{9,6} = 1.00$$

$$A_{9;4} = \begin{pmatrix} 0.4300 & 0.4480 \\ 0.1430 & 0.0040 \\ 0.4270 & 0.5480 \end{pmatrix}, A_{9;5} = \begin{pmatrix} 0.3570 & 0.1530 \\ 0.4880 & 0.4430 \\ 0.1540 & 0.4040 \end{pmatrix}, A_{9;6} = \begin{pmatrix} 0.5260 & 0.4240 \\ 0.2360 & 0.4750 \\ 0.2380 & 0.1010 \end{pmatrix}$$

In the traditional algorithm-based system, the exact value of $Pr\{E | B_{1,1}\}$ is $7.939915 \times 10^{-2}$, and it cost 71ms.

Meanwhile, assign the burn-in steps value as $b = 300$, window width as $\omega = 200$, error rate expected as $\epsilon = 10^{-3}$, and confidence $95\% (\Delta = 5\%$ and $C = 2)$. Then set the layer threshold $IG_{layer} = 2$ and $X$ threshold $IG_x = 6$. Repeat the algorithm 6 times, then the sampling result is Table 5, and the convergence process is shown in Figure 12.

| Attempts | Loop count | Time consumed | Exact result | Sampling result | Error Ratio |
|---|---|---|---|---|---|
| 1st Running | 1282 | 0.375S | 7.9399e-02 | 7.9457e-02 | 0.073% |
| 2nd Running | 1217 | 0.363S | 7.9399e-02 | 8.0137e-02 | 0.930% |
| 3rd Running | 1876 | 0.550S | 7.9399e-02 | 7.9314e-02 | -0.107% |
| 4th Running | 1967 | 0.572S | 7.9399e-02 | 7.9077e-02 | -0.406% |
| 5th Running | 1105 | 0.319S | 7.9399e-02 | 7.8971e-02 | -0.539% |
| 6th Running | 1536 | 0.458S | 7.9399e-02 | 7.9730e-02 | 0.417% |

Table 5:Sampling result for example in Figure 11

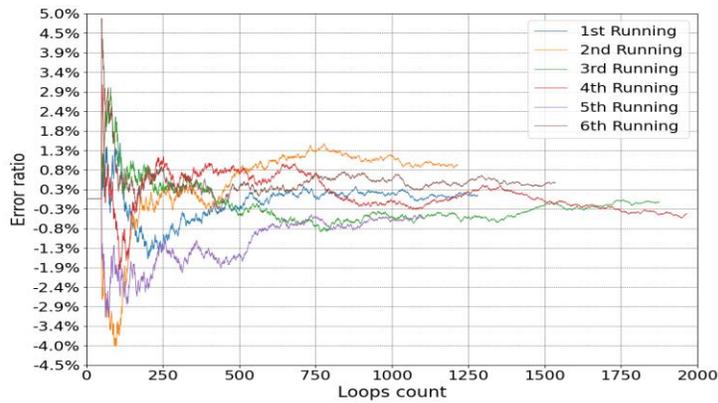

Figure 12:Convergence process for example in Figure 11

As it is shown in Table 5, all of 6 tests converge after about 1100 loops and costs about 400ms, which verifies the stability of algorithm's halting rule. In addition, all of 6 tests end with an error ration within −0.6%~1%, and the fluctuation is slight after the convergence.

Because the scale of the graph is small, sampling method cost more time than the traditional one, but this example managed to verify the accuracy of the new sampling algorithm.

## 4.2. Full joined n × n model

In order to provide an intuitive presentation to the performance advantages over traditional algorithm in expanding process, several full joined models with $n \times n$ $X$ variables are tested in this section, where $n$ grows from 2 to 10.

The model is presented before in Figure 6. Assign the directed-arcs with random relation matrix $A$, with the constraint that sum of each column is 1, then assign all causal relations of $F$-type events as $r_{ij} = 1$.

For more details, the $2 \times 2$ example in Figure 13 is presented to show the initialization of data.

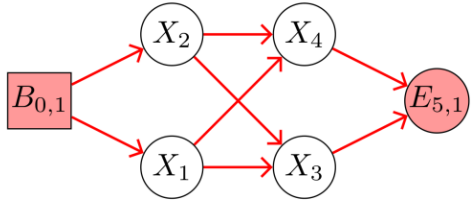

Figure 13: $2 \times 2$ full joined model

As it is shown in the model, the root cause is $B_{0,1}$ and the evidence is $E = E_{5,1}$. The relation matrixes are randomly generated:

$$r_{1,0} = 1.00, r_{2,0} = 1.00$$

$$A_{1;0} = \begin{pmatrix} 0.5769 & 0.0698 & 0.3239 \\ 0.2969 & 0.6470 & 0.3678 \\ 0.1262 & 0.2831 & 0.3083 \end{pmatrix}, A_{2;0} = \begin{pmatrix} 0.1749 & 0.3294 & 0.3849 \\ 0.2320 & 0.1924 & 0.2738 \\ 0.5932 & 0.4782 & 0.3414 \end{pmatrix}$$

$$r_{3,1} = 1.00, r_{3,2} = 1.00$$

$$A_{3;1} = \begin{pmatrix} 0.0423 & 0.7340 & 0.3815 \\ 0.2950 & 0.1439 & 0.3696 \\ 0.6628 & 0.1221 & 0.2489 \end{pmatrix}, A_{3;2} = \begin{pmatrix} 0.3079 & 0.3693 & 0.4806 \\ 0.1673 & 0.3585 & 0.4309 \\ 0.5248 & 0.2722 & 0.0886 \end{pmatrix}$$

$$r_{4,1} = 1.00, r_{4,2} = 1.00$$

$$A_{4;1} = \begin{pmatrix} 0.6019 & 0.7720 & 0.1163 \\ 0.1055 & 0.2084 & 0.6770 \\ 0.2926 & 0.0197 & 0.2067 \end{pmatrix}, A_{4;2} = \begin{pmatrix} 0.3966 & 0.1356 & 0.3562 \\ 0.1306 & 0.5973 & 0.3558 \\ 0.4729 & 0.2671 & 0.2879 \end{pmatrix}$$

$$r_{5,3} = 1.00, r_{5,4} = 1.00$$

$$A_{5;3} = \begin{pmatrix} 0.6376 & 0.2926 & 0.4112 \\ 0.2876 & 0.4266 & 0.0886 \\ 0.0748 & 0.2808 & 0.5002 \end{pmatrix}, A_{5;4} = \begin{pmatrix} 0.3786 & 0.5072 & 0.2200 \\ 0.5989 & 0.3092 & 0.3565 \\ 0.0225 & 0.1835 & 0.4235 \end{pmatrix}.$$

The parameters in sampling process are assigned same as the Sec.4.1, where burn-in loops $b = 300$, window width $\omega = 200$, $\epsilon = 10^{-3}$ and $\Delta \approx 5\%$ (C = 2). Because this example has only on evidence, the cutting off scheme for absorbing process is not required. Therefore, the traditional algorithm-based program is also simplified by removing the absorbing logic in this test.

With all mentioned above, full-joined models with n = 2, 3, 4...10 are tested by both traditional numerical and stochastic sampling algorithms for comparison.

Note that because the time consumption of traditional method would reach 292.3s when $n = 8$, the time required for traditional method would be too large when $n$ is larger. Therefore, only sampling method is tested for $n = 9, 10$.

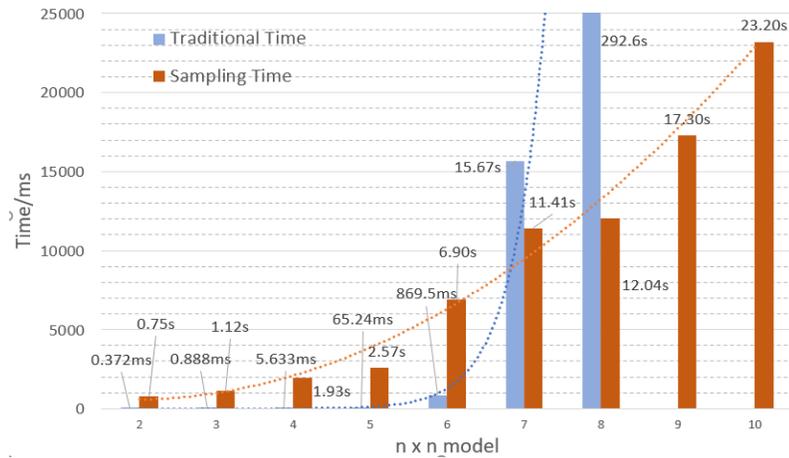

Figure 14:Time consumption of the algorithm

Figure 14 shows the time required for the two algorithms. Apparently, traditional method spends less time than sampling one when n < 6. Meanwhile, time consumption for traditional algorithm raises rapidly when $n \geq 7$, which shows the trend of typical combinatorial explosion. In comparison, the increment of time required by sampling method is flat.

Approximate curves for traditional algorithm present as an exponential function, while the curve for sampling one display as a polynomial function, which is consistent with the derivation of complexity in previous sections.

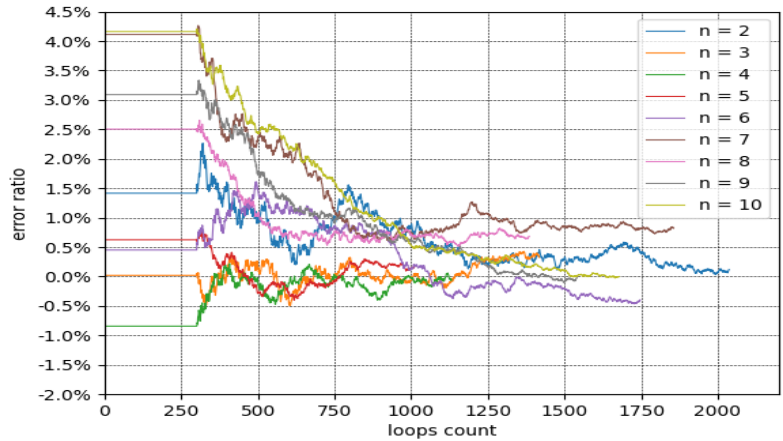

Figure 15:Convergence process for n × n model

| n | Loop count | Traditional result | Sampling result | Error ratio |
|---|---|---|---|---|
| 2 | 2037 | 0.365 | 0.364 | 0.176% |
| 3 | 1417 | 0.231 | 0.232 | 0.327% |
| 4 | 1131 | 0.152 | 0.152 | -0.061% |
| 5 | 1010 | 0.191 | 0.191 | 0.175% |
| 6 | 1747 | 0.119 | 0.119 | -0.397% |
| 7 | 1856 | 0.082 | 0.082 | 0.823% |
| 8 | 1385 | 0.101 | 0.102 | 0.685% |
| 9 | 1541 | - | 0.061 | 0.000% |
| 10 | 1678 | - | 0.062 | 0.000% |

Table 6:Accuracy of the sampling algorithm

Figure 15 shows the convergence process of the 9 cases. The situation is similar to Figure 12: after the peak at beginning, curves shocks slightly in a tiny interval around the final result. More detailed data about loop counts and the error ratio is listed in the Table 6. All tests end with error less than 1%, which verifies the accuracy for sampling algorithm as an approximate method.

Note that loop counts for all examples are about 1000 ∼ 2000 regardless of the n value, which verifies the discussion about loop counts in Sec.3.3.1. Moreover, as the path from root cause to the leaves node are denoted by domain engineers in practical use, most of models are less than 10 layers, so that thousands of loops will be enough for practical uses.

## 4.3. An example for Viral Hepatitis B

After the discussions about idealized models, an example adapted from the practical model for viral hepatitis B is presented for further verification. For more accuracy, the result of traditional algorithm is calculated by a production edition program deployed in hospitals, instead of an experimental one used in previous examples. This example has large scale of both expanding process and absorbing process.

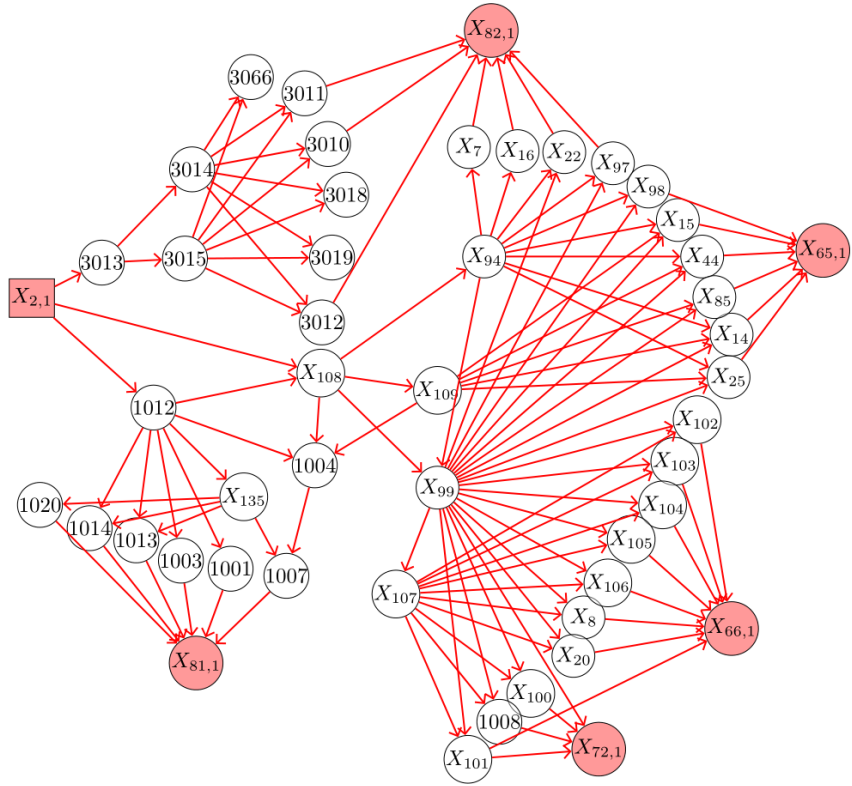

Figure 16:A DUCG model for Viral Hepatitis B

This example is developed from the graph built in [20]. As it is shown in the Figure 14, the model consists of 49 nodes and 109 directed arcs. The scale of this example is smaller than the full-joined model, but the relationship between evidences nodes are more complex.

Definition of nodes are listed in Table 7. The evidences are $E = X_{65,1}X_{66,1}X_{72,1}X_{81,1}X_{82,1}$, which means the patient is suffering these symptoms.

Relation matrices for parts of *F-type* events are listed below, some of these parameters are assigned by doctors, and normalization are executed to fill the missing parameters. Most of causal relations $r_{ij}$ are 1, due to a rapid and flexible construction demanded by doctors.

$$r_{3013,2} = 1.00, r_{1012,2} = 1.00, r_{108,2} = 1.00$$

$$A_{3013;2} = \begin{pmatrix} 0.8250 & 0.2350 \\ 0.1750 & 0.7650 \end{pmatrix}, A_{1012;2} = \begin{pmatrix} 0.6970 & 0.8660 \\ 0.3030 & 0.1340 \end{pmatrix}, A_{108;2} = \begin{pmatrix} 0.2060 & 0.1340 \\ 0.7940 & 0.8660 \end{pmatrix}$$

$$r_{81,1020} = 1.00, r_{81,1014} = 1.00, r_{81,1013} = 1.00$$

$$A_{81;1020} = \begin{pmatrix} 0.3990 & 0.6870 \\ 0.6010 & 0.3130 \end{pmatrix}, A_{81;1014} = \begin{pmatrix} 0.6450 & 0.3840 \\ 0.3550 & 0.6160 \end{pmatrix}, A_{81;1013} = \begin{pmatrix} 0.6570 & 0.4770 \\ 0.3430 & 0.5230 \end{pmatrix}$$

$$r_{81,1003} = 1.00, r_{81,1001} = 1.00, r_{81,1007} = 1.00$$

$$A_{81;1003} = \begin{pmatrix} 0.0660 & 0.6270 \\ 0.9340 & 0.3730 \end{pmatrix}, A_{81;1001} = \begin{pmatrix} 0.1380 & 0.4650 \\ 0.8620 & 0.5350 \end{pmatrix}, A_{81;1007} = \begin{pmatrix} 0.3890 & 0.1800 \\ 0.6110 & 0.8200 \end{pmatrix}$$

$$r_{65,25} = 1.00, r_{65,14} = 1.00, r_{65,85} = 1.00$$

$$A_{65;25} = \begin{pmatrix} 0.0490 & 0.4550 \\ 0.9510 & 0.5450 \end{pmatrix}, A_{65;14} = \begin{pmatrix} 0.2840 & 0.1640 \\ 0.7160 & 0.8360 \end{pmatrix}, A_{65;85} = \begin{pmatrix} 0.4040 & 0.3990 \\ 0.5960 & 0.6010 \end{pmatrix}$$

| Node Id | Node Definition | Node Id | Node Definition |
|---|---|---|---|
| $B_2$ | Viral hepatitis B | $X_{103}$ | Liver and kidney syndrome |
| $X_7$ | Fever | $X_{104}$ | Hepatopulmonary syndrome |
| $X_8$ | Jaundice | $X_{105}$ | ARDS |
| $X_{14}$ | Diarrhea | $X_{106}$ | Secondary infection |
| $X_{15}$ | Bloating | $X_{107}$ | Subacute liver failure |
| $X_{16}$ | Darker urine | $X_{108}$ | Symptoms of hepatitis B virus |
| $X_{20}$ | Hepatic encephalopathy | $X_{109}$ | Chronic hepatitis B symptoms |
| $X_{22}$ | Urticaria | $X_{135}$ | Liver cirrhosis |
| $X_{25}$ | Constipation | $X_{1001}$ | Large liver |
| $X_{44}$ | Fatigue | $X_{1003}$ | Splenomegaly |
| $X_{65}$ | Headache | $X_{1004}$ | Liver swelling and pain |
| $X_{66}$ | Dizziness | $X_{1007}$ | Ascites |
| $X_{72}$ | Trembling | $X_{1008}$ | Bleeding tendency |
| $X_{81}$ | Shortness of breath after activity | $X_{1012}$ | Signs of hepatitis B virus |
| $X_{82}$ | Lower limb edema | $X_{1013}$ | Spider nevus |
| $X_{85}$ | Joint pain | $X_{1014}$ | Liver palm |
| $X_{94}$ | Acute hepatitis symptoms | $X_{1020}$ | Abdominal Varicose Veins |
| $X_{97}$ | Vasculitis | $X_{3010}$ | Serum total bilirubin |
| $X_{98}$ | glomerulus nephritis | $X_{3011}$ | Elevated ALT |
| $X_{99}$ | Acute liver failure | $X_{3012}$ | AST increased |
| $X_{100}$ | Significant fatigue | $X_{3013}$ | Laboratory test for viral hepatitis B |
| $X_{101}$ | Irritability | $X_{3014}$ | Laboratory tests for acute hepatitis B |
| $X_{102}$ | Brain edema | $X_{3015}$ | Laboratory tests for chronic hepatitis B |
| $X_{3019}$ | Anti-HBc-IgG | $X_{3018}$ | Anti-HBc-IgM |
| $X_{3066}$ | Mildly elevated alpha-fetoprotein | | |

Table 7: Definition of nodes in Figure 16

The result for traditional method is $2.8773 \times 10^{-2}$ and time consumed is 105.9s.

Because the relationship between evidences are complex, the sampling parameters is adapted for better performance. The burn-in step is assigned b = 200, windows width is $\omega = 100$, $\epsilon = 0.3\%$, $\Delta = 5\%$ (C = 2), while IG $_{layer}$ = 2 and $IG_{len}$ = 5. The test is repeated 5 times for a stable assessment. The results are listed in Table 8 and the convergence path is illustrated in Figure 17.

| Attempts | Loop count | Time consumed | Exact result | Sampling result | Error Ratio |
|---|---|---|---|---|---|
| 1st Running | 639 | 31.595s | 2.8773e-02 | 2.8046e-02 | -2.525% |
| 2nd Running | 548 | 26.710s | 2.8773e-02 | 2.9146e-02 | 1.297% |
| 3rd Running | 799 | 38.515s | 2.8773e-02 | 2.8789e-02 | 0.054% |
| 4th Running | 571 | 26.563s | 2.8773e-02 | 2.9550e-02 | 2.701% |
| 5th Running | 571 | 26.959s | 2.8773e-02 | 2.8089e-02 | -2.377% |

Table 8: Sampling result for example in Figure 16

All tests converge within 800 cycles, and the final results are in the interval $2.8046 \times 10^{-2} \sim 2.9550 \times 10^{-2}$. The average of time consumption of 5 running is 29.8s,

Despite the error ratio increases a little to 2.5% due to the change of parameters, the sampling algorithm cost only $1/3$ time of the traditional one, and the error ratio is still less than 5% which meets the accuracy demanded.

It can be figured out that the estimate scheme for absorbing processes cost lots of time in each cycle, so the count of cycles has to be reduced. A better estimation algorithm can be helpful, which will be addressed as future work.

In conclusion, this new sampling algorithm provides a highly efficient and accurate method for inference on large scale DUCG.

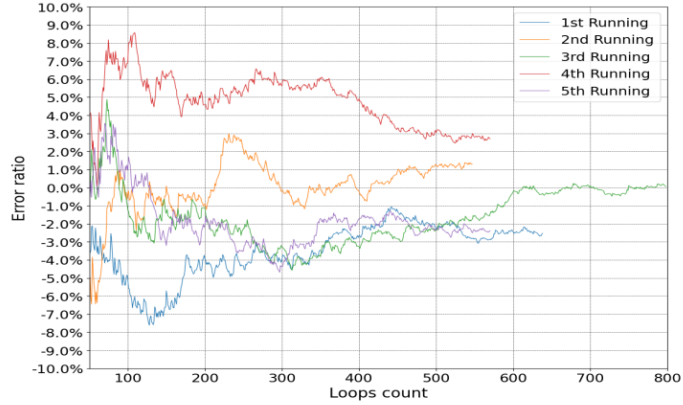

Figure 17: Convergence for sampling in Figure 16

## 5. Conclusion and future work

In this paper, the stochastic simulation algorithm for DUCG inference is proposed for the expanding and logic operation combination explosion problems. Then a theoretical analysis on algorithm's time complexity is demonstrated based on two typical examples.

For the expanding problem, the time complexity is reduced from $O(n^{n+1})$ to $O\left(n^3 \left(C\delta/\epsilon\mu\right)^2\right)$ with sampling algorithm, while $(C\delta/\epsilon\mu)^2$ is less than $10^4$ in practical cases. For the logic operation problem, the complexity is reduced from $O(n^2 e^n)$ to $O(IG^2 e^{IG})$, where IG is a hyper parameter which is less than 0.5n in most cases.

Two ideal examples are presented to verify the accuracy and derivations about time complexity, and both cases convergence with the error rate less than 1%. Then a practical model about Viral Hepatitis B is tested. The sampling algorithm runs three times faster than the traditional one on this example, and the error rate is less than 2.5%.

In conclusion, the algorithm provides a high-efficient and accurate method for large scale DUCG models, which will help DUCG based system to handle with more complex situations, and will allow domain engineers to add more not observed variables for more accurate description.

Nevertheless, due to the limit of time, some deficiencies of theory could not be ignored, i.e., firstly, burn-in parameters $b$ and window width parameters $\omega$ is assigned based on experience, so that a little deviation between actual and theoretical result still remains; secondly, estimation for logic operation is based on hyper parameters $IG_{layer}$ and $IG_{len}$, so this scheme may fail in some edge situations; at last, as sampling method cost more time in small scale models, a practical selector is required for choosing between sampling and traditional algorithm.

Researches are planned to address these shortages in future works. And more real cases in practical use will be presented in the following papers.

# Appendix A: Recursive inference method for DUCG

With the chain rule shown in Eqn.(8), the detailed scheme of recursive method is Algorithm 3.

---

**Algorithm 3:** Recursive algorithm for DUCG inference

**Input:** Count of layers as $l_{max}$

1 **Initial:**
2 Assign each $X_i \in E$ with its layer parameter $l_i$, which is the arcs count of the shortest path between $X_i$ and the $B$ in sub-DUCG.
3 Denote $E(k) = \bigcup_{l_i=k} X_i$
4 **Inference:**
5 $k = l_max$
6 $Exp = 1$;
7 **for** $k > 1$ **do**
8     $Exp_k = 1$
9     **for** $X_i \in E(k)$ **do**
10       $Exp_i = expand\ X_i\ to\ \bigcup_{l_j} X_j$
11       $Exp_k = Exp_k \bigcap Exp_i$
12     **end**
13     Substitute the observed state and numerical parameter of $\bigcup_{X_i \in E(k)} X_i$ into Exp, so that
      /* $Exp = Pr\{\bigcup_{n>k} E(n) | B \bigcup_{n<=k} E(n)\}$ */
14     $Exp = expand\ Exp\ to\ \bigcup_{l_j} X_j$
      /* $Exp_k = Pr\{E(k) | B \bigcup_{n<k} E(n)\}$, therefore $Pr\{\bigcup_{n>(k-1)} E(n) | B \bigcup_{n<=(k-1)} E(n)\}$ */
15     $Exp = Exp \bigcap Exp_k$
16     $k = k - 1$
17 **end**
18 Substitute the wanted state of $B$ to Exp and get the final result

$$Pr\{\bigcup_{k}^{l_{max}} E(k) | B\}$$

---

With the scheme, the expanding expression can be broken into several shorter expressions, which can help to reduce the impact of combination explosion.

Figure 18 is an example to show the scheme for recursive algorithm[5]. The root cause is $B_{1,1}$, and the evidences are $X_2 = 1$, $X_4 = 1$ and $X_6 = 1$. The red color indicates non-0 state, and no color means its state is unknown. For convenience, all $r$ parameters are assigned 1.

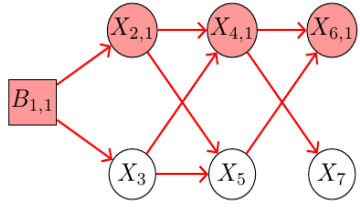

Figure 18:Example of recursive algorithm

The aim of inference is to get:

$$Pr\{E \mid B_{1,1}\}$$
$$= Pr\{X_{2,1}X_{4,1}X_{6,1} \mid B_{1,1}\}$$

In traditional algorithm, the expanding process is Eqn.(27):

$$
\begin{aligned}
&Pr\{X_{2,1}X_{4,1}X_{6,1} \mid B_{1,1}\} \\
&= Pr\{F_{2,1;1,1}B_{1,1} \cdot (F_{4,1;2,1}F_{2,1;1,1} + F_{4,1;3}F_{3;1,1})B_{1,1} \cdot [F_{6,1;4,1}X_{4,1} + F_{6,1;5}(F_{5;2,1}X_{2,1} + F_{5;3,1}F_{3,1;1,1}B_{1,1})] \mid B_{1,1}\} \\
&= \frac{Pr\{F_{2,1;1,1}B_{1,1} \cdot (F_{4,1;2,1}F_{2,1;1,1} + F_{4,1;3}F_{3;1,1})B_{1,1}}{\cdot F_{6,1;4,1}(F_{4,1;2,1}F_{2,1;1,1} + F_{4,1;3}F_{3;1,1})B_{1,1} + F_{6,1;5}(F_{5;2,1}F_{2,1;1,1}B_{1,1} + F_{5;3,1}F_{3,1;1,1}B_{1,1})] \mid B_{1,1}\}} \\
&= Pr\{F_{2,1;1,1} \cdot (F_{4,1;2,1}F_{2,1;1,1} + F_{4,1;3}F_{3;1,1}) \cdot [F_{6,1;4,1}(F_{4,1;2,1}F_{2,1;1,1} + F_{4,1;3}F_{3;1,1}) \\
&\quad + F_{6,1;5}(F_{5;2,1}F_{2,1;1,1} + F_{5;3,1}F_{3,1;1,1})]B_{1,1} \mid B_{1,1}\} \\
&= F_{2,1;1,1}(F_{4,1;2,1}F_{2,1;1,1} + F_{4,1;3}F_{3;1,1}) \cdot [F_{6,1;4,1}(F_{4,1;2,1}F_{2,1;1,1} + F_{4,1;3}F_{3;1,1}) \\
&\quad + F_{6,1;5}(F_{5;2,1}F_{2,1;1,1} + F_{5;3,1}F_{3,1;1,1})]
\end{aligned}
\quad (27)
$$

Because the $F-$ type parameter are regarded as event, this equation will be simplified by the absorb rules in [1] and then be calculated for a numerical result. The longest expression in Eqn.(27) contains 15 $F-$ type events.

As a comparison, the process of the recursive algorithm is presented below.

Because all evidences are located in different layers, the expression can be reformed as:

$$Pr\{X_{2,1}X_{4,1}X_{6,1} \mid B_{1,1}\}$$
$$= Pr\{X_{2,1} \mid B_{1,1}\} \cdot Pr\{X_{4,1} \mid B_{1,1}X_{2,1}\} \cdot Pr\{X_{6,1} \mid X_{4,1}X_{2,1}B_{1,1}\}$$

Then, $X_{2,1}$ can be expanded as:

$$X_{2,1} = F_{2,1;1,1}B_{1,1}$$

Expand expression for $X_{4,1}$ is:

$$X_{4,1} = F_{4,1;2,1}X_{2,1} + F_{4,1;3}F_{3;1,1}B_{1,1}$$

After that, $X_{6,1}$ is:

$$\begin{aligned} X_{6,1} &= F_{6,1;4,1}X_{4,1} + F_{6,1;5}X_5 \\ &= F_{6,1;4,1}X_{4,1} + F_{6,1;5}(F_{5;2,1}X_{2,1} + F_{5;3,1}F_{3,1;1,1}B_{1,1}) \end{aligned}$$

Therefore, the result will be Eqn.(28)

$$Pr\{E \mid B_{1,1}\} = F_{2,1;1,1} \cdot (F_{4,1;2,1} + F_{4,1;3}F_{3;1,1}) \cdot [F_{6,1;4,1} + F_{6,1;5}(F_{5;2,1} + F_{5;3,1}F_{3,1;1,1})] \quad (28)$$

This expression is equal to the result of recursive one. Note that it contains only 9 $F-type$ events, which means less complexity of space and time.

# References

[1] Q. Zhang, "Dynamic Uncertain Causality Graph for Knowledge Representation and Reasoning: Discrete {DAG} Cases," *J. Comput. Sci. Technol.*, vol. 27, no. 1, pp. 1–23, 2012, doi: 10.1007/s11390-012-1202-7.

[2] Q. Zhang, C. Dong, Y. Cui, and Z. Yang, "Dynamic Uncertain Causality Graph for Knowledge Representation and Probabilistic Reasoning: Statistics Base, Matrix, and Application," *IEEE Trans. Neural Networks Learn. Syst.*, vol. 25, no. 4, pp. 645–663, 2014, doi: 10.1109/TNNLS.2013.2279320.

[3] Q. Zhang, "Dynamic Uncertain Causality Graph for Knowledge Representation and Probabilistic Reasoning: Directed Cyclic Graph and Joint Probability Distribution," *IEEE Trans. Neural Networks Learn. Syst.*, vol. 26, no. 7, pp. 1503–1517, 2015, doi: 10.1109/tnnls.2015.2402162.

[4] Z. Qin and Z. Zhan, "Dynamic Uncertain Causality Graph Applied to Dynamic Fault Diagnoses and Predictions With Negative Feedbacks," *IEEE Trans. Reliab.*, vol. 65, no. 2, pp. 1030–1044, 2016.

[5] Q. Zhang and Q. Yao, "Recursive algorithm in DUCG," *Intern. Tech. Rep. (in Chinese), available by Req. from Corresp. author.*, 2017.

[6] Q. Zhang, X. Bu, M. Zhang, Z. Zhang, and J. Hu, "Dynamic uncertain causality graph for computer-aided general clinical diagnoses with nasal obstruction as an illustration," *Artif. Intell. Rev.*, Jul. 2020, doi: 10.1007/s10462-020-09871-0.

[7] Q. Zhang, "Probabilistic reasoning based on dynamic causality trees/diagrams," *Reliab. Eng. Syst. Saf.*, vol. 46, no. 3, pp. 209–220, 1994, doi: 10.1016/0951-8320(94)90114-7.

[8] C. Dong, Z. Zhou, and Q. Zhang, "Cubic Dynamic Uncertain Causality Graph: A New Methodology for Modeling and Reasoning About Complex Faults With Negative Feedbacks," *IEEE Trans. Reliab.*, vol. 67, no. 3, pp. 920–932, 2018, doi: 10.1109/tr.2018.2822479.

[9] Z. Qin and Z. Zhan, "Extended method for constructing intelligent system processing uncertain causal relationship type information." 2018.

[10] Y. Zhao, F. Di Maio, E. Zio, Q. Zhang, C.-L. Dong, and J.-Y. Zhang, "Optimization of a dynamic uncertain causality graph for fault diagnosis in nuclear power plant," *Nucl. Sci. Tech.*, vol. 28, no. 3, p. 34, 2017.

[11] Y. Quanying, Z. Qin, L. Peng, Y. Ping, Z. Ma, and W. Xiaochen, "Application of dynamic uncertain causality graph in spacecraft fault diagnosis: Multi-conditions," *AIP Conf. Proc.*, vol. 1834, no. 1, p. 30012, 2017, doi: 10.1063/1.4981577.

[12] G. F. Cooper, "The computational complexity of probabilistic inference using bayesian belief networks," *Artif. Intell.*, vol. 42, no. 2–3, pp. 393–405, 1990, doi: 10.1016/0004-3702(90)90060-d.

[13] M. HENRION, "Propagating Uncertainty in Bayesian Networks by Probabilistic Logic Sampling," in *Uncertainty in Artificial Intelligence*, Elsevier, 1988, pp. 149–163.

[14] H. Chai, J. Lei, and M. Fang, "Estimating bayesian networks parameters using {EM} and Gibbs sampling," *Procedia Comput. Sci.*, vol. 111, pp. 160–166, 2017, doi: 10.1016/j.procs.2017.06.023.

[15] R. Fung and K.-C. Chang, "Weighing and Integrating Evidence for Stochastic Simulation in Bayesian Networks," in *Uncertainty in Artificial Intelligence*, Elsevier, 1990, pp. 209–219.

[16] P. Dagum and E. Horvitz, "A Bayesian analysis of simulation algorithms for inference in belief networks," *Networks*, vol. 23, no. 5, pp. 499–516, 1993, doi: 10.1002/net.3230230506.

[17] D. Hongchen and Z. Qin, "Cubic DUCG (Dynamical Uncertain Causality Graph) Recursive Algorithm and Its Implementation in Nuclear Power Plant Fault Diagnoses," Tsinghua University, Beijing, 2018.

[18] R. M. Karp, M. Luby, and N. Madras, "Monte-Carlo approximation algorithms for enumeration problems," *J. Algorithms*, vol. 10, no. 3, pp. 429–448, 1989, doi: 10.1016/0196-6774(89)90038-2.

[19] M. K. Cowles and B. P. Carlin, "Markov Chain Monte Carlo Convergence Diagnostics: A Comparative Review," *J. Am. Stat. Assoc.*, vol. 91, no. 434, pp. 883–904, 1996.

[20] S. Hao, S. Geng, L. Fan, J. Chen, Q. Zhang, and L. Li, "Intelligent diagnosis of jaundice with dynamic uncertain causality graph model," *J. Zhejiang Univ. B*, vol. 18, no. 5, pp. 393–401, 2017, doi: 10.1631/jzus.b1600273.